\documentclass{article}

\usepackage[main,final]{neurips_2026}

\usepackage{algorithmic}
\usepackage{algorithm}
\usepackage{xcolor} 
\usepackage{colortbl}
\usepackage{multirow}
\usepackage{multicol}
\usepackage{makecell}
\usepackage{booktabs}
\usepackage{graphicx}
\usepackage{amsmath}
\usepackage{pifont}
\usepackage{fvextra}

\usepackage[normalem]{ulem}

\newcommand{\aycite}[1]{(\citeauthor{#1}, \citeyear{#1})}

\usepackage[utf8]{inputenc}
\usepackage[T1]{fontenc}
\usepackage{hyperref}
\usepackage{url}
\usepackage{amsfonts}
\usepackage{nicefrac}
\usepackage{microtype}

\setcitestyle{numbers,square}

\title{AutoVLA-D: Learn a Trajectory-Geometric Condition from Reasoning for Autonomous Driving VLA}

\author{
Yuguang Yang$^{1,3}$\thanks{Equal contribution:\{guangbuaa, chencanyu, tanzhewen\}@buaa.edu.cn}\quad
Zhewen Tan$^{8}$ \footnotemark[1] \quad
Canyu Chen$^{2,3}$ \footnotemark[1] \quad\\
Linlin Yang$^{5}$ \quad
Cheng Chi$^{7}$ \quad
Chunyang Liu$^{4}$ \quad 
Kehua Sheng$^{4}$ \quad
Jinyu Yang $^{9}$ \quad \\
Baochang Zhang$^{6}$\thanks{Project Lead}  \quad 
Bo Zhang$^{4}$ \thanks{Corresponding author: \texttt{xbcao@buaa.edu.cn}, \texttt{zhangbo@didiglobal.com}, \texttt{wangyan@air.tsinghua.edu.cn}} \quad
Yan Wang $^{3}$ \footnotemark[3]   \quad
Xianbin Cao $^{1}$ \footnotemark[3]  \quad 
\\
$^{1}$School of Electronic and Information Engineering, Beihang University \\
$^{2}$National College for Excellent Engineers, Beihang University\\
$^{3}$Institute for AI Industry Research (AIR), Tsinghua University \\ 
$^{4}$Didi \\
$^{5}$State Key Laboratory of Media Convergence and Communication, \\
Communication University of China \\
$^{6}$ Hangzhou Innovation Research Institute, Beihang University \\
$^{7}$School of Information, Renmin University of China\\
$^{8}$School of Computer Science and Engineering, Beihang University  \\
$^{9}$Harbin Institute of Technology, Shenzhen
}

\begin{document}
\maketitle

\begin{abstract}
Autoregressive VLA planning usually predicts continuous trajectories as the final output, but driving instructions often describe actions rather than ego states, which weakens direct conditioning on the final continuous trajectory. We address this gap with \textit{TrajCond}, which introduces a discrete trajectory-geometric condition for VLA planning. Specifically, we represent TrajCond as a 1D-Brohan differential action trajectory and learn it with a two-stage training framework before final continuous trajectory prediction. With the simultaneous autoregressive generation of discrete and numeric trajectories, we name our method \textbf{AutoVLA-D} (Double). Experiments on NAVSIM navtest, nuScenes, and NeuroNCAP show that discrete trajectories are weaker as direct outputs but more effective as TrajCond, consistently improving final continuous trajectory prediction across closed-loop and open-loop driving benchmarks. Code is available at \url{https://github.com/ShivaTeam/AutoVLA-D}.
\end{abstract}

\section{Introduction}
\label{sec:introduction}

Autoregressive Vision-Language-Action (VLA) planning has become a practical Multimodal Large Language Model (MLLM)-native paradigm for autonomous driving. Given camera observations and driving instructions, a VLA planner follows the next-token interface of modern MLLMs: it conditions on visual-language inputs, produces intermediate driving reasoning, and then decodes the future ego trajectory~\cite{rowe2025poutine}. Recent methods have demonstrated strong closed-loop planning capability with this input--reasoning--trajectory generation process~\cite{zhou2025autovla,chen2026devil,luo2025adathinkdrive,luo2026elfvla,li2025drivevlaw0}.

Despite this progress, existing VLA planners can also be distinguished by the trajectory form used for the final ego plan. One line of work models the future ego states, including the relative $x$-axis position, $y$-axis position, and heading in the current ego frame, over the next few seconds as numeric text waypoint tokens~\cite{luo2025adathinkdrive,luo2026elfvla,chen2026devil}, which we call \emph{continuous trajectory} in this paper; while another line, \textit{i.e.}, \emph{discrete trajectory}, represents the step-wise ego motion, or state transition, as discrete action tokens~\cite{wu2024smart,li2025drivevlaw0,zhou2025autovla}. These two trajectory forms have complementary strengths and limitations. Action tokens introduce reconstruction error when discretizing continuous trajectories, and our experiments over multiple discretization strategies, \textit{e.g.}, K-Means, K-Disks~\cite{zhou2025autovla,wu2024smart,philiontrajeglish}, and Brohan-style discretization~\cite{zitkovich2023rt}, show that directly regressing discrete trajectories is consistently weaker than directly regressing continuous trajectories. However, driving instructions often describe actions such as yielding, keeping, or turning, and such local motion changes are naturally aligned with discrete trajectory tokens because each action token describes a local state transition. This creates a formulation gap when instructions are mapped directly to the final continuous trajectory.

To address this, we propose \textbf{AutoVLA-D} (Double), a two-stage VLA planner that uses a discrete trajectory as a \emph{Trajectory-geometric Condition} (TrajCond) before regressing the final continuous trajectory. This approach features three core innovations:

\begin{itemize}

\item \textbf{Discrete trajectory as trajectory-geometric condition.}
TrajCond first predicts a discrete trajectory and then uses its {geometric-semantic} as a condition for final continuous prediction. This keeps the final output in the stronger continuous space while using the discrete condition to transmit action-oriented information.

\item \textbf{Action tokens with 1D-Brohan discretization.}
We construct TrajCond from differential ego motions and discretize the $x$, $y$, and yaw differentials with separate Brohan-style codebooks. Compared with shared-codebook clustering tokenizers, such as K-Means and K-Disks~\cite{zhou2025autovla,wu2024smart,philiontrajeglish}, this design gives the strongest tested TrajCond result and the lowest tested reconstruction error.

\item \textbf{Two-stage alignment training.}
After constructing Brohan TrajCond by discretizing the ground-truth trajectory, the newly constructed action vocabulary has no VLA-task-related initialization. Stage 1 therefore warms up TrajCond's semantics by aligning observations and instructions with the action-token sequence that encodes TrajCond. Stage 2 uses the decoded TrajCond and the available reasoning context to regress the final continuous trajectory. This separates condition learning from final trajectory prediction.

\end{itemize}

We evaluate AutoVLA-D on NAVSIM navtest, nuScenes, and NeuroNCAP. AutoVLA-D achieves state-of-the-art results across these settings, including 92.0 PDMS and 86.8 EPDMS on NAVSIM navtest. Beyond benchmark comparisons, our experiments are organized to verify how and why trajectory forms matter for driving VLAs. In TrajCond Ablation and Progressive Analysis, we examine the staged performance gains of AutoVLA-D and validate the contribution of its training and reasoning components. In TrajCond Form Analysis, we compare continuous trajectories with different action-token forms, showing that discrete trajectories are weaker as final autoregressive outputs but become more effective when used as TrajCond for final continuous prediction. In TrajCond Semantic Analysis, we qualitatively and quantitatively verify whether TrajCond absorbs action-oriented planning semantics in addition to providing geometric guidance. Together, these experiments suggest that in autoregressive VLAs, discrete trajectories used as TrajCond can encode instruction semantics while preserving trajectory geometry, thereby reducing the formulation gap between instructions and continuous trajectories and improving VLA planning performance.

\section{Related Work}
\label{sec:related}

\noindent\textbf{Autonomous Driving VLA.}
Autonomous driving planning first developed through end-to-end systems that map sensor observations to future trajectories, including ST-P3~\cite{hu2022st}, VAD~\cite{jiang2023vad}, UniAD~\cite{hu2023planning}, DiffusionDrive~\cite{liao2025diffusiondrive}, SparseDrive~\cite{sun2025sparsedrive}, and BridgeAD~\cite{zhang2025bridging}. With the emergence of vision-language models (VLMs), recent VLA methods exploit visual-semantic knowledge for driving understanding, reasoning, and planning. They follow two patterns: planner-style methods, including ReCogDrive~\cite{li2025recogdrive}, ImagiDrive-S~\cite{li2025imagidrive}, DriveVLA-W0~\cite{li2025drivevlaw0}, and FLARE~\cite{xie2026flare}, use the VLM as a feature extractor or semantic reasoner with an external planner or action head; autoregressive methods, including EMMA~\cite{hwang2024emma}, OpenEMMA~\cite{xing2025openemma}, OpenDriveVLA~\cite{zhou2025opendrivevla}, Impromptu VLA~\cite{chi2025impromptu}, AutoVLA~\cite{zhou2025autovla}, AdaThinkDrive~\cite{luo2025adathinkdrive}, Curious-VLA~\cite{chen2026devil}, and ELF-VLA~\cite{luo2026elfvla}, predict waypoint or action-token sequences through the next-token interface. Unlike autoregressive VLAs that directly decode the final trajectory after reasoning, AutoVLA-D inserts a learned trajectory-geometric condition before final continuous trajectory generation.

\noindent\textbf{Trajectory Form in Autoregressive VLA.}
Autoregressive VLA planners represent the future ego trajectory over three degrees of freedom: $x$-axis position, $y$-axis position, and heading. Existing trajectory forms can be grouped into two types. Continuous-trajectory methods, including Impromptu VLA~\cite{chi2025impromptu}, AdaThinkDrive~\cite{luo2025adathinkdrive}, Curious-VLA~\cite{chen2026devil}, and ELF-VLA~\cite{luo2026elfvla}, serialize each future ego state as numeric text waypoint tokens; among them, Curious-VLA~\cite{chen2026devil} further improves performance with step-wise trajectory normalization. Discrete-trajectory methods instead serialize each local ego-state transition as an action token and recover the future trajectory by temporal integration. AutoVLA~\cite{zhou2025autovla} builds K-Means/K-Disks-style action codebooks whose tokens represent 3-DoF local motions~\cite{wu2024smart,philiontrajeglish}, while DriveVLA-W0~\cite{li2025drivevlaw0} uses frequency-based FAST tokenization~\cite{pertsch2025fast}. AutoVLA-D differs by using discrete trajectories as trajectory-geometric conditions rather than final outputs, and adopts a 1D-Brohan form where each action token represents a 1-DoF local motion.

\begin{figure}
    \centering
    \includegraphics[width=1\linewidth]{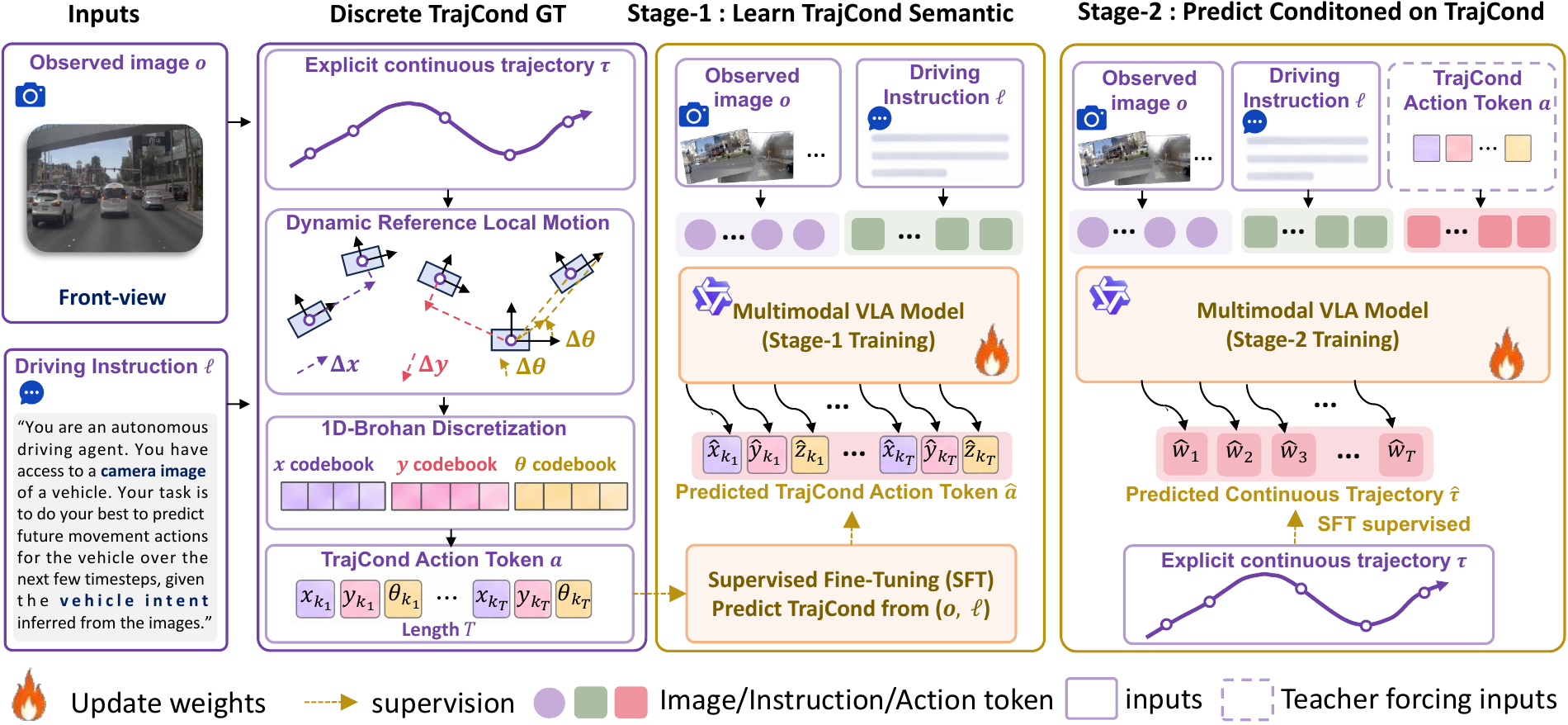}
    \caption{Pipeline of AutoVLA-D.}
    \label{fig:pipeline}
\vspace{-15pt}
\end{figure}
\section{Method}

\label{sec:method}

\subsection{Preliminary \& Formulation}
\label{sec:formulation}

\noindent\textbf{Task formulation.}
Given observed images $\mathbf{o}$ and a linguistic user instruction $\ell$, a conventional VLA planner directly predicts the final future trajectory:
\begin{equation}
\label{eq:vla-direct}
\tau = f_{\text{VLA}}(\mathbf{o}, \ell),
\end{equation}
In this paper, the final prediction target $\tau$ is always a continuous trajectory, represented for autoregressive decoding as the normalized text-waypoint sequence $\{\tilde{\tau}_t\}_{t=1}^{T}$. This direct formulation therefore asks the model to map language instructions to the final continuous trajectory autoregressively.

As shown in Fig.~\ref{fig:pipeline}, \textbf{AutoVLA-D}, introduces an explicit intermediate \emph{trajectory-geometric condition}. Let $\mathbf{a}=\{a_t\}_{t=1}^{T}$ denote the action-token sequence that encodes this condition. Unlike the final output trajectory, TrajCond is discrete. Instead of mapping $\ell$ directly to the final continuous trajectory $\tau$, we train the same VLA model through two supervised stages:
\begin{equation}
\label{eq:two-stage}
\mathbf{a}=f_{\text{VLA}}^{(1)}(\mathbf{o},\ell),
\qquad
\tau=f_{\text{VLA}}^{(2)}(\mathbf{o},\ell,\mathbf{a}),
\quad
f_{\text{VLA}}^{(2)} \leftarrow f_{\text{VLA}}^{(1)}.
\end{equation}
where superscripts indicate the training stage rather than separate architectures, and $f_{\text{VLA}}^{(2)} \leftarrow f_{\text{VLA}}^{(1)}$ means that the Stage-2 model is initialized from the Stage-1 model and then further trained. Both stages use the MLLM's autoregressive token decoder: Stage 1 predicts TrajCond action tokens from $(\mathbf{o},\ell)$, and Stage 2 predicts final continuous waypoint tokens from $(\mathbf{o},\ell,\mathbf{a})$.

\noindent\textbf{Continuous trajectory preliminary.}
Given observed images $\mathbf{o}$ and a linguistic user instruction $\ell$, the model predicts a future ego trajectory $\tau=\{(x_t,y_t,\theta_t)\}_{t=1}^{T}$. Following Curious-VLA~\cite{chen2026devil}, the final continuous trajectory is serialized as normalized text waypoints. Concretely, letting $\tau_t=(x_t,y_t,\theta_t)$ denote the trajectory state at step $t$, we use the step-wise normalized representation
\begin{equation}
\tilde{\tau}_t = \frac{\tau_t - \mu_t}{\sigma_t}, \quad
\hat{\tau}_t = \hat{\tilde{\tau}}_t \sigma_t + \mu_t,
\end{equation}
where $(\mu_t,\sigma_t)$ are per-step statistics estimated from the training set. We use \emph{TrajCond} to bridge action-oriented instructions and continuous ego-state outputs.

\noindent\textbf{Context preliminary.}
\label{sec:context-preliminary}
For all datasets, $\ell$ denotes the linguistic user instruction. For NAVSIM, we additionally use Qwen3-VL-235B to annotate Chain-of-Thought (CoT) reasoning from the front-view image, high-level intent, past ego trajectory, and expert future trajectory. Following prior VLA reasoning practice~\cite{rowe2025poutine}, the annotation contains three fields: critical objects and conditions, a natural-language explanation of the expert trajectory, and speed/command meta-behavior labels. The CoT sequence can be viewed as an extension term associated with $\ell$ in Eq.~\ref{eq:two-stage}, but it is not provided as input; instead, the model is trained to regress it as additional language supervision. The full annotation prompt and the dataset-specific Stage-1/Stage-2 user prompts are provided in Supplement Sec.~\ref{sec:supp-system-prompt}. For nuScenes and NeuroNCAP, no CoT annotation is used.
For fair comparison, $\mathbf{o}$ uses a single front-view image on NAVSIM and nuScenes, and left-front-right three view images on NeuroNCAP.

\subsection{Trajectory-Geometric Condition via Action Tokens}
\label{sec:trajcond}

A useful TrajCond should be predictable with discrete tokens while retaining enough geometry to guide continuous planning. We therefore build TrajCond from trajectory differentials, which preserve local motion changes while admitting a compact discrete vocabulary.

\noindent\textbf{Action tokenizer.}
The trajectory $\tau=\{(x_t,y_t,\theta_t)\}_{t=1}^{T}$ is expressed in the ego frame at $t=0$. Unlike previous works that rely on a static coordinate frame~\cite{zhou2025autovla, li2025drivevlaw0}, to describe the motion from step $t-1$ to $t$, we use the local frame at step $t-1$ and compute the differential motion with a standard planar rotation transform~\cite{lynch2017modern}. Supplement Fig.~\ref{fig:local-frame-differential} illustrates this local-frame differential construction:
\begin{equation}
\label{eq:local_frame_delta}
\begin{aligned}
\delta_t &= (\Delta x_t, \Delta y_t, \Delta \theta_t), \quad
\Delta \theta_t = \theta_t-\theta_{t-1},\\
\left[\begin{smallmatrix}
\Delta x_t\\
\Delta y_t
\end{smallmatrix}\right]
&=
\mathbf{R}(-\theta_{t-1})
\left[\begin{smallmatrix}
x_t-x_{t-1}\\
y_t-y_{t-1}
\end{smallmatrix}\right],\quad
\mathbf{R}(\theta)=
\left[\begin{smallmatrix}
\cos\theta & -\sin\theta\\
\sin\theta & \cos\theta
\end{smallmatrix}\right].
\end{aligned}
\end{equation}
Let $\Delta \tau=\{\delta_t\}_{t=1}^{T}$ denote the differential trajectory. Following previous action-token practice~\cite{zhou2025autovla,wu2024smart,philiontrajeglish}, we extract differential actions from training trajectories and build an action-template set. Existing action-template extraction methods commonly use K-Means or K-Disks to cluster the joint 3-DoF motion vector $(\Delta x,\Delta y,\Delta \theta)$ into one shared codebook of size $Q$. In our analysis, this is the \emph{3D shared-codebook} form because each time step is represented by one action token that encodes a 3-DoF local motion.

In contrast, we find in Sec.~\ref{sec:trajectory_representation} that clustering-based K-Means/K-Disks tokenizers produce a clear train-test reconstruction-error gap. Our final setting therefore follows the linear per-dimension discretization strategy used in RT-2~\cite{zitkovich2023rt}. For each action dimension $i \in \{1,2,3\}$ corresponding to $x$, $y$, and $\theta$, we compute the 1st and 99th quantiles on the training set and uniformly divide that interval into $Q_i$ bins. The total action-token codebook size is $Q=\sum_{i=1}^{3} Q_i$. This yields three codebooks $\mathbf{x}$, $\mathbf{y}$, and $\boldsymbol{\theta}$, one per motion dimension, namely the \emph{1D-Brohan} form where each individual action token represents a 1-DoF local motion and the three tokens at a time step jointly describe a 3-DoF local transition.

\noindent\textbf{Trajectory discretization.}
Following~\cite{philiontrajeglish,zhou2025autovla,wu2024smart}, we iteratively find the token with minimum corner distance to the next state. In our final 1D-Brohan setting, a candidate tuple $\mathbf{k}=(k_1,k_2,k_3)$ selects one bin center from each action codebook:
\begin{equation}
\label{eq:brohan_candidate}
\bar{\delta}(\mathbf{k})=(x_{k_1},y_{k_2},\theta_{k_3}),
\quad k_i \in \{1,\ldots,Q_i\}.
\end{equation}
Starting from the current ego pose $\hat{\tau}_0=(0,0,0)$, we render each candidate from the previously tokenized pose $\hat{\tau}_{t-1}$:
\begin{equation}
\label{eq:action_recon}
\left[\begin{smallmatrix}\bar{x}_t(\mathbf{k})\\ \bar{y}_t(\mathbf{k})\end{smallmatrix}\right]
=
\left[\begin{smallmatrix}\hat{x}_{t-1}\\ \hat{y}_{t-1}\end{smallmatrix}\right]
+\mathbf{R}(\hat{\theta}_{t-1})
\left[\begin{smallmatrix}x_{k_1}\\ y_{k_2}\end{smallmatrix}\right],
\quad
\bar{\theta}_t(\mathbf{k})=\hat{\theta}_{t-1}+\theta_{k_3}.
\end{equation}
The best candidate tuple for step $t$ is selected by comparing the rendered candidate state with the ground-truth next state:
\begin{equation}
\label{eq:action_token}
\mathbf{k}_t^\ast=\arg\min_{\mathbf{k}} d_{\mathrm{corner}}\bigl((\bar{x}_t(\mathbf{k}),\bar{y}_t(\mathbf{k}),\bar{\theta}_t(\mathbf{k})),(x_t,y_t,\theta_t)\bigr),
\end{equation}
where $d_{\mathrm{corner}}$ is the mean distance between the ordered box corners. The action-token tuple is $a_t=\mathbf{k}_t^\ast$. Iterating this process yields the full discrete action sequence $\mathbf{a}=\{a_t\}_{t=1}^{T}$. Since the three 1D codebooks correspond to different DoFs, exhaustive tuple search would require $Q_i^3$ candidates at each step. We therefore first take the four tokens with the smallest absolute per-dimension residuals for each DoF, enumerate the resulting $4\times4\times4$ candidate tuples, and choose the tuple with the smallest corner error. This design converts a continuous trajectory into a sequence of local action components; Supplement Sec.~\ref{sec:supp-traj-differential} provides the visualization and pseudocode for the full process.

\subsection{AutoVLA-D Training \& Inference}
\label{sec:training}

Here we detail the training and inference process described in the task formulation of Sec.~\ref{sec:formulation}.

\noindent\textbf{Stage-1 Inference \& Training.}
The newly constructed action vocabulary has no VLA-task-related initialization, so Stage 1 warms up TrajCond's semantics by taking multimodal vision and language context $(\mathbf{o},\ell)$ as input and outputting the action-token sequence $\mathbf{a}$. We train this stage with Supervised Fine-Tuning (SFT) using the autoregressive token loss:
\begin{equation}
\label{eq:stage1}
\mathcal{L}_{\text{stage1}} = -\sum_{t=1}^{T} \log P(a_t \mid \mathbf{o}, \ell, a_{<t}).
\end{equation}
Stage 1 is not used to produce the final trajectory directly; it learns TrajCond as the trajectory-geometric condition that guides Stage 2.

\noindent\textbf{Stage-2 Inference \& Training.}
This stage model learns to use TrajCond $\mathbf{a}$ together with visual and linguistic instruction context $(\mathbf{o},\ell)$ to regress the target continuous trajectory. Let $\mathbf{w}=\{w_t\}_{t=1}^{T}$ denote the token sequence corresponding to the target continuous trajectory. Stage 2 takes the same multimodal context $(\mathbf{o},\ell)$ as input and follows the output order $\mathbf{a}\rightarrow\mathbf{w}$. We train this stage with SFT, and the objective combines the action-token loss and the final waypoint-token loss:
\begin{equation}
\label{eq:stage2}
\mathcal{L}_{\text{stage2}} =
-\sum_{t=1}^{T} \log P(a_t \mid \mathbf{o}, \ell, a_{<t})
-\sum_{t=1}^{T} \log P(w_t \mid \mathbf{o}, \ell, \mathbf{a}, w_{<t}).
\end{equation}
Stage 2 optimizes the full ordered output sequence that contains both TrajCond action tokens and final trajectory tokens. As shown in the dashed box in Fig.~\ref{fig:pipeline}, we adopt the teacher-forcing strategy, where action tokens are supervised by ground truth trajectory during training, while inference is performed in a fully autoregressive manner that TrajCond firstly predicts TrajCond action trajectory, then predicts the final continuous trajectory.

\noindent\textbf{Reasoning \& RFT.}
For NAVSIM, we additionally use CoT reasoning, as described in the context preliminary of Sec.~\ref{sec:context-preliminary}. Let $\mathbf{r}$ denote the CoT reasoning tokens; the output order becomes $\mathbf{r}\rightarrow\mathbf{a}$ in Stage 1 and $\mathbf{a}\rightarrow\mathbf{r}\rightarrow\mathbf{w}$ in Stage 2. Following Curious-VLA~\cite{chen2026devil}, we add a third training stage with the output format $\mathbf{a}\rightarrow\mathbf{r}\rightarrow\mathbf{w}$, and use GRPO~\cite{shao2024deepseekmath} to perform Reinforcement Fine-Tuning (RFT). For fair comparison, nuScenes and NeuroNCAP use only Stage-1/2 SFT training without CoT reasoning.

As supported by the trajectory-representation analysis in Fig.~\ref{fig:discrete-error}, directly predicting discrete trajectories is weaker than predicting the final continuous trajectory, but using the discrete trajectory as TrajCond strengthens the final planner. The two-stage design therefore preserves the strength of the continuous output space while exploiting the alignment advantage of TrajCond.

\section{Experiments}
\label{sec:experiments}

In Sec.~\ref{sec:main_results}, we report performance on NAVSIM navtest, nuScenes, and NeuroNCAP. In Sec.~\ref{sec:progressive}, we conduct TrajCond ablation and progressive analysis. In Sec.~\ref{sec:trajectory_representation}, we compare different TrajCond forms and validate the compatibility of TrajCond with different discrete trajectory tokenizers. In Sec.~\ref{sec:trajcond_semantic_validation}, we conduct semantic analyses of TrajCond.

\subsection{Experimental Setup}
\label{sec:setup}

\noindent\textbf{Datasets and benchmark settings.}
We report results on \textbf{NAVSIM navtest}~\cite{dauner2024navsim,chen2026devil}, a large-scale non-reactive closed-loop simulation benchmark, \textbf{nuScenes}~\cite{caesar2020nuscenes}, an open-loop real-world driving benchmark, and \textbf{NeuroNCAP}~\cite{ljungbergh2024neuroncap}, a photorealistic closed-loop simulator built on nuScenes. For fair comparison, we follow ReCogDrive~\cite{li2025recogdrive} to report NAVSIM navtest, follow Impromptu VLA~\cite{chi2025impromptu} to report nuScenes, and follow ImagiDrive~\cite{li2025imagidrive} to report NeuroNCAP. More benchmark details are provided in Supplement Sec.~\ref{sec:supp-benchmark-details}.

\noindent\textbf{Training setup.}
For NAVSIM and the representation ablations, we use Qwen3-VL-4B unless otherwise specified. For nuScenes and NeuroNCAP, we report both Qwen2.5-VL and Qwen3-VL variants. Training consists of SFT followed by optional RFT. We implement SFT with LLaMA-Factory and RFT with EasyR1. We optimize with AdamW~\cite{loshchilov2017decoupled}, weight decay $0.05$, warmup ratio $0.10$, cosine learning-rate scheduler, and global batch size $32$. Stage 1 and Stage 2 each run for 3 epochs with learning rate $4\times 10^{-5}$. {For RFT, we perform GRPO training for 160 steps with an actor batch size of 512, broadly consistent with Curious-VLA~\aycite{chen2026devil}.} The model-context setup for NeuroNCAP, NAVSIM, and nuScenes is provided in context preliminary of  Sec.~\ref{sec:context-preliminary}.

\noindent\textbf{Trajectory formulation.}
Unless otherwise specified, \emph{continuous trajectory} denotes the default final trajectory form expressed as normalized text waypoints, \emph{TrajCond} denotes discrete trajectory encoded by 1D-Brohan $Q{=}768$ codebooks, where each DoF's codebook has $Q_i{=}256$ bins. We choose $Q{=}768$ after comparing different codebook-size settings in Sec.~\ref{sec:trajectory_representation}.

\subsection{Benchmark Comparisons}
\label{sec:main_results}

\begin{table*}[t]
\centering
\scriptsize
\setlength{\tabcolsep}{5pt}
\renewcommand{\arraystretch}{0.8}
\caption{{NAVSIM navtest PDMS results under the public metric setting.} $^\dagger$ denotes results reproduced by us. $^*$ denotes concurrent work.}
\label{tab:main_pdms}
\resizebox{0.90\textwidth}{!}{%
\begin{tabular}{lccccccc}
\toprule
Method & Base Model & NC$\uparrow$ & DAC$\uparrow$ & EP$\uparrow$ & TTC$\uparrow$ & C$\uparrow$ & PDMS$\uparrow$ \\
\midrule
ImagiDrive-S & InternVL2.5-4B & 98.6 & 96.2 & 80.5 & 94.5 & \textbf{100.0} & 87.4 \\
ReCogDrive & InternVL3-8B & 98.2 & 97.8 & 83.5 & 95.2 & \textbf{100.0} & 89.6 \\
ReCogDrive$^\dagger$ & Qwen3-VL-4B & 98.2 & 97.7 & 86.0 & 97.7 & \textbf{100.0} & 90.7 \\
DriveVLA-W0 & Emu3-8B & \underline{98.7} & \textbf{99.1} & 83.3 & 95.3 & 99.3 & 90.2 \\
AutoVLA & Qwen2.5-VL-3B & 98.4 & 95.6 & 81.9 & \textbf{98.0} & \underline{99.9} & 89.1 \\
AdaThinkDrive & InternVL3-8B & 98.4 & 97.8 & 84.4 & 95.2 & \textbf{100.0} & 90.3 \\
Curious-VLA & Qwen2.5-VL-3B & 98.4 & 96.9 & \underline{88.5} & \underline{97.9} & 98.1 & 90.3 \\
Curious-VLA$^\dagger$ & Qwen3-VL-4B & 98.5 & 96.7 & 87.3 & \textbf{98.0} & 98.3 & 89.7 \\
FLARE$^*$ & Qwen3-VL-4B & 98.5 & \underline{98.4} & 86.0 & 96.0 & \textbf{100.0} & \underline{91.4} \\
\midrule
ours & Qwen3-VL-4B & \textbf{98.8} & 97.7 & \textbf{89.8} & 97.7 & 99.8 & \textbf{92.0} \\
\bottomrule
\end{tabular}
}
\vspace{-10pt}
\end{table*}

\noindent\textbf{NAVSIM navtest PDMS results.}
Table~\ref{tab:main_pdms} reports the NAVSIM navtest comparison. The compared methods include VLA-planner methods ImagiDrive-S~\aycite{li2025imagidrive}, ReCogDrive~\aycite{li2025recogdrive}, and DriveVLA-W0~\aycite{li2025drivevlaw0}; and VLA-token methods AutoVLA~\aycite{zhou2025autovla}, AdaThinkDrive~\aycite{luo2025adathinkdrive}, Curious-VLA~\aycite{chen2026devil}, and FLARE~\aycite{xie2026flare}. Additional baselines are provided in Supplement Table~\ref{tab:supp-navsim-pdms-e2e}. Our NAVSIM setting uses front-view input, reason data, and optional GRPO~\cite{shao2024deepseekmath}. Experiment results demonstrated that AutoVLA-D reaches 92.0 PDMS and exceeds the reported scores of prior VLA planners.

\begin{table*}[t]
\centering
\scriptsize
\setlength{\tabcolsep}{2.4pt}
\renewcommand{\arraystretch}{1.00}
\caption{{NAVSIM navtest EPDMS results following the official EPDMS protocol.} AutoVLA-D uses the same 1-camera setting as the PDMS experiment and reports the completed EPDMS evaluation. $^\dagger$ denotes results reproduced by us. $^*$ denotes concurrent work. {EPDMS
denotes the bug-fixed EPDMS metric released by the official repository of NAVSIM v2.}}
\label{tab:main_epdms}
\resizebox{\textwidth}{!}{%
\begin{tabular}{llcccccccccc}
\toprule
Method & Base Model & NC$\uparrow$ & DAC$\uparrow$ & DDC$\uparrow$ & TLC$\uparrow$ & EP$\uparrow$ & TTC$\uparrow$ & LK$\uparrow$ & C$\uparrow$ & EC$\uparrow$ & EPDMS$\uparrow$  \\
\midrule
ReCogDrive & InternVL3-8B & 98.3 & 95.2 & \textbf{99.5} & \underline{99.8} & 87.1 & 97.5 & 96.6 & \underline{98.3} & 86.5 & 83.6 \\
DiffusionDrive & -- & 98.2 & 95.9 & \underline{99.4} & \underline{99.8} & 87.5 & 97.3 & \underline{96.8} & \underline{98.3} & \textbf{87.7} & 84.5 \\
Curious-VLA & Qwen2.5-VL-3B & 98.4 & 96.9 & 99.2 & \underline{99.8} & \underline{88.5} & 97.9 & \textbf{96.9} & 98.1 & 81.5 & 85.3 \\
Curious-VLA$^\dagger$ & Qwen3-VL-4B & \underline{98.5} & 96.7 & 99.37 & \textbf{99.9} & 87.3 & 98.0 & 96.5 & \underline{98.3} & 83.0 & 84.9 \\
DriveVLA-W0 & Emu3-8B & \underline{98.5} & \textbf{99.1} & 98.0 & 99.7 & 86.4 & \underline{98.1} & 93.2 & 97.9 & 58.9 & 86.1 \\
FLARE$^*$ & Qwen3-VL-4B & \textbf{98.8} & 96.6 & \textbf{99.5} & \textbf{99.9} & 87.9 & \textbf{98.2} & 95.8 & \textbf{98.4} & \underline{87.5} & \underline{86.3} \\
\midrule
ours & Qwen3-VL-4B & \textbf{98.8}& \underline{97.7}& 98.8& \underline{99.8}& \textbf{91.2}& \textbf{98.2}& \textbf{96.9}& 98.2& 80.2& \textbf{86.8}\\
\bottomrule
\end{tabular}
}\vspace{-10pt}
\end{table*}

\noindent\textbf{NAVSIM navtest EPDMS results.}
Table~\ref{tab:main_epdms} reports the same benchmark under the stricter EPDMS metric set. The compared methods include ReCogDrive~\aycite{li2025recogdrive}, DiffusionDrive~\aycite{liao2025diffusiondrive}, DriveVLA-W0~\aycite{li2025drivevlaw0}, Curious-VLA~\aycite{chen2026devil}, and FLARE~\aycite{xie2026flare}; additional E2E baselines are provided in Supplement Table~\ref{tab:supp-navsim-epdms-e2e}. Under this setting, AutoVLA-D reaches 86.8 EPDMS and improves over the reported Curious-VLA~\aycite{chen2026devil} score of 85.3.

\begin{table*}[t]
\centering
\begin{minipage}[t]{0.48\textwidth}
\vspace{0pt}
\centering
\scriptsize
\setlength{\tabcolsep}{1.8pt}
\renewcommand{\arraystretch}{1.12}
\caption{{Open-loop comparison on nuScenes.} We report L2 error and collision rate (Col.). Gray rows denote own ablation w/o TrajCond (TC).}
\label{tab:nuscenes_l2}
\resizebox{\linewidth}{!}{%
\begin{tabular}{llcccc}
\toprule
\multirow{2}{*}{Method} & \multirow{2}{*}{Base Model} & \multicolumn{2}{c}{ST-P3 metrics} & \multicolumn{2}{c}{UniAD metrics} \\
\cmidrule(lr){3-4} \cmidrule(lr){5-6}
 & & L2 $\downarrow$ & Col. $\downarrow$ & L2 $\downarrow$ & Col. $\downarrow$ \\
\midrule
ST-P3 & -- & 2.11 & 0.71 & -- & -- \\
VAD & -- & 0.37 & 0.14 & -- & -- \\
UniAD & -- & 0.69 & 0.12 & 1.03 & 0.31 \\
EMMA & Gemini & 0.32 & -- & -- & -- \\
OpenEMMA & Qwen2-VL-7B & 2.81 & -- & -- & -- \\
OpenDriveVLA & Qwen2.5-VL-3B & 0.33 & \underline{0.10} & 0.67 & \underline{0.30} \\
AutoVLA & Qwen2.5-VL-3B & 0.48 & 0.13 & 0.86 & 0.35 \\
Impromptu VLA & Qwen2.5-VL-3B & 0.33 & 0.13 & 0.67 & 0.38 \\
\midrule
Ours & Qwen2.5-VL-3B & \underline{0.30} & 0.10 & \underline{0.60} & 0.35 \\
\rowcolor{gray!12} Ours w/o TC & Qwen3-VL-4B & 0.30 & 0.11 & 0.60 & 0.34 \\
\rowcolor{gray!12} Ours & Qwen3-VL-4B & \textbf{0.30} & \textbf{0.09} & \textbf{0.60}  & \textbf{0.29} \\
\bottomrule
\end{tabular}
}
\vspace{-10pt}
\end{minipage}
\hfill
\begin{minipage}[t]{0.49\textwidth}
\vspace{0pt}
\centering
\scriptsize
\setlength{\tabcolsep}{1.5pt}
\renewcommand{\arraystretch}{1.47}
\caption{{Closed-loop comparison on NeuroNCAP.} Gray rows denote own ablation w/o TrajCond (TC).}
\label{tab:neuroncap}
\resizebox{\linewidth}{!}{%
\begin{tabular}{llcccccccc}
\toprule
\multirow{2}{*}{Method} & \multirow{2}{*}{Base Model} & \multicolumn{4}{c}{NeuroNCAP Score $\uparrow$} & \multicolumn{4}{c}{Collision rate $\downarrow$} \\
\cmidrule(lr){3-6} \cmidrule(lr){7-10}
& & Avg. & Stat. & Frontal & Side & Avg. & Stat. & Frontal & Side \\
\midrule
UniAD & -- & 0.73 & 0.84 & 0.10 & 1.26 & 0.89 & 0.88 & 0.98 & 0.80 \\
VAD & -- & 0.66 & 0.47 & 0.04 & 1.45 & 0.93 & 0.96 & 1.00 & 0.82 \\
SparseDrive & -- & 0.92 & -- & -- & -- & 0.94 & -- & -- & -- \\
BridgeAD-S & -- & 1.52 & -- & -- & -- & 0.76 & -- & -- & -- \\
BridgeAD-B & -- & 1.60 & -- & -- & -- & 0.73 & -- & -- & -- \\
ImagiDrive-S & InternVL2.5-4B & \underline{3.49} & \textbf{4.15} & 2.45 & \textbf{3.88} & 0.45 & \underline{0.34} & 0.74 & \textbf{0.27} \\
Impromptu VLA & Qwen2.5-VL-3B & 2.15 & 1.77 & 2.31 & 2.10 & 0.66 & 0.70 & 0.59 & 0.65 \\
\midrule
\rowcolor{gray!12} Ours w/o TC & Qwen2.5-VL-3B & 2.76 & 2.95 & 2.98 & 2.35 & 0.49 & 0.43 & 0.48 & 0.56 \\
Ours & Qwen2.5-VL-3B & \textbf{3.63} & \underline{4.02} & \underline{3.43} & \underline{3.44} & \textbf{0.36} & \textbf{0.26} & \underline{0.44} & \underline{0.36} \\
\rowcolor{gray!12} Ours w/o TC & Qwen3-VL-4B & 1.66 & 1.81 & 1.06 & 1.66 & 0.68 & 0.69 & 0.82 & 0.68 \\
\rowcolor{gray!12} Ours & Qwen3-VL-4B & 3.23 & 3.17 & \textbf{3.68} & 2.86 & \underline{0.39} & 0.41 & \textbf{0.32} & 0.44 \\
\bottomrule
\end{tabular}
}
\vspace{-15pt}
\end{minipage}
\end{table*}

\noindent\textbf{nuScenes open-loop comparison.}
Table~\ref{tab:nuscenes_l2} reports the nuScenes open-loop comparison against ST-P3~\aycite{hu2022st}, VAD~\aycite{jiang2023vad}, UniAD~\aycite{hu2023planning}, EMMA~\aycite{hwang2024emma}, OpenEMMA~\aycite{xing2025openemma}, OpenDriveVLA~\aycite{zhou2025opendrivevla}, AutoVLA~\aycite{zhou2025autovla}, and Impromptu VLA~\aycite{chi2025impromptu}. The Qwen2.5-VL row is the primary setting, while the gray Qwen3-VL rows isolate backbone and TrajCond ablations. Notably, the L2 error on nuScenes has become nearly saturated at approximately 0.30 m across existing methods, while collision rate provides a more discriminative measure.

\noindent\textbf{NeuroNCAP closed-loop comparison.}
Table~\ref{tab:neuroncap} reports the NeuroNCAP closed-loop comparison using NeuroNCAP Score and collision rate. The compared methods include UniAD~\aycite{hu2023planning}, VAD~\aycite{jiang2023vad}, BridgeAD~\aycite{zhang2025bridging}, ImagiDrive-S~\aycite{li2025imagidrive}, and Impromptu VLA~\aycite{chi2025impromptu}, together with the additional E2E rows shown in the table. The Qwen2.5-VL AutoVLA-D row is the primary setting, and the gray rows show that removing TrajCond or switching backbone weakens the result.

\subsection{TrajCond Ablation \& Progressive Analysis}
\label{sec:progressive}

\noindent\textbf{Progressive analysis on NAVSIM.}
\begin{table}[t]
\centering
\caption{{Progressive performance improvement on NAVSIM with Qwen3-VL-4B.} We abbreviate Reason, Continuous, Brohan, and epochs as R, C, B, and ep., respectively. The right panel reports the corresponding EPDMS/PDMS scores.}
\label{tab:progressive}
\newsavebox{\progressivetablebox}
\newlength{\progressivepanelheight}
\sbox{\progressivetablebox}{%
\resizebox{0.54\linewidth}{!}{%
\renewcommand{\arraystretch}{1.18}%
\begin{tabular}{l|c|c|c}
\toprule
\textbf{Method} & \textbf{Stage 1} & \textbf{Stage 2} & \textbf{RFT} \\
\midrule
\multicolumn{4}{l}{\textit{SFT-only Results w/ Reason (R)}} \\
\midrule
Continuous (C) & R $\rightarrow$ C (6 ep.) & -- & -- \\
Continuous TrajCond & R $\rightarrow$ C (3 ep.) & C $\rightarrow$ R $\rightarrow$ C (3 ep.) & -- \\
Brohan (B) & R $\rightarrow$ B (6 ep.) & -- & -- \\
Brohan TrajCond (1) & -- & B $\rightarrow$ R $\rightarrow$ C (6 ep.) & -- \\
Brohan TrajCond & R $\rightarrow$ B (3 ep.) & B $\rightarrow$ R $\rightarrow$ C (3 ep.) & -- \\
\midrule
\multicolumn{4}{l}{\textit{SFT + RFT (GRPO) Results w/ Reason (R)}} \\
\midrule
Continuous + RFT & R $\rightarrow$ C (6 ep.) & -- & \checkmark \\
Brohan TrajCond + RFT & R $\rightarrow$ B (3 ep.) & B $\rightarrow$ R $\rightarrow$ C (3 ep.) & \checkmark \\
\bottomrule
\end{tabular}
}
}
\setlength{\progressivepanelheight}{\dimexpr\ht\progressivetablebox+\dp\progressivetablebox\relax}
\begin{minipage}[t]{0.55\linewidth}
\vspace{0pt}
\noindent\usebox{\progressivetablebox}
\end{minipage}%
\hspace{-0.025\linewidth}%
\begin{minipage}[t]{0.47\linewidth}
\vspace{0pt}
\centering
\noindent\includegraphics[height=\progressivepanelheight]{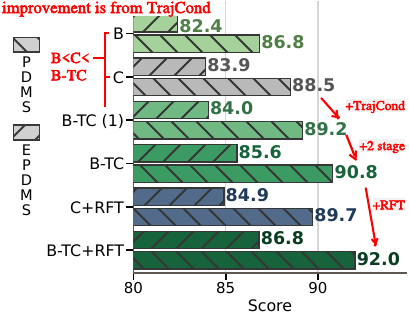}
\end{minipage}

\vspace{-20pt}
\end{table}
\noindent Table~\ref{tab:progressive} traces the improvement path from continuous trajectory regression to the full AutoVLA-D recipe. A one-stage Brohan TrajCond already improves over the direct continuous baseline (\texttt{C} < \texttt{B-TC(1)}), and separating the process into two stages improves further (\texttt{B-TC(1)} < \texttt{B-TC}) because TrajCond should first absorb action-oriented semantics before being used as the condition for final continuous trajectory prediction. This progression also clarifies that Brohan is weaker as the direct output target (\texttt{B}<\texttt{C}) but stronger as an intermediate condition (\texttt{B}<\texttt{C}<\texttt{B-TC}), suggesting that the gain of AutoVLA-D does not come from distilling discrete trajectories, but from using discrete trajectories as a condition.

\noindent\textbf{TrajCond ablation on nuScenes/NeuroNCAP.}
The completed comparisons further validate AutoVLA-D beyond NAVSIM. In the SFT-only nuScenes setting, TrajCond improves open-loop planning over the w/o-TC variant under both ST-P3 and UniAD metrics. On NeuroNCAP, TrajCond also improves closed-loop safety and reduces collisions. This supports that TrajCond is not a NAVSIM-only artifact.

\begin{figure}[t]
    \centering
    \includegraphics[width=1\linewidth]{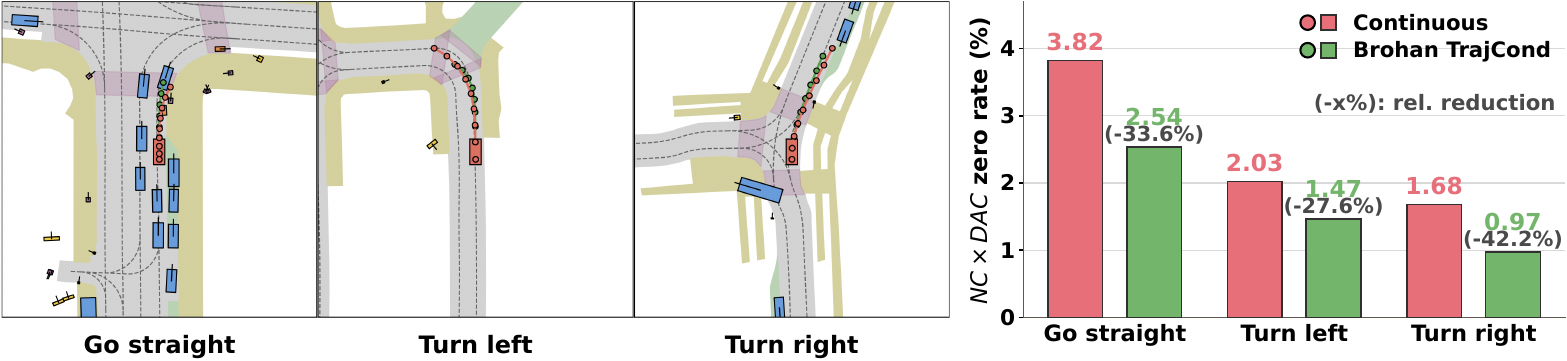}
    \caption{{Collision examples and failed $NC{\times}DAC$ product rates on NAVSIM navtest.} The left three panels show representative collision-related scenes for the three high-level intents, and the right panel reports the failed $NC{\times}DAC$ product rate for Continuous and Brohan TrajCond. Full NC/DAC/Product counts are provided in Supplement Table~\ref{tab:supp-nc-dac-stat}.}
\vspace{-10pt}
    \label{fig:nc-dac-product-reduction}
\end{figure}
\noindent\textbf{Attribution analysis.}
The closed-loop results indicate that TrajCond improves collision-related behavior most clearly. Our technical route is also close to Curious-VLA~\cite{chen2026devil} except for TrajCond, and compared with this baseline, the main gain comes from the reduction of collision rate. We therefore further analyze NAVSIM collision-related metrics in Fig.~\ref{fig:nc-dac-product-reduction}. The figure focuses on No Collisions ($\mathrm{NC}$) and Drivable Area Compliance ($\mathrm{DAC}$), and shows representative cases where the high-level meta-behavior is already correct, but TrajCond produces more accurate local control and avoids collision-prone motion. The ablation rows in Tables~\ref{tab:nuscenes_l2} and~\ref{tab:neuroncap} show a consistent trend on nuScenes and NeuroNCAP, suggesting that TrajCond improves safety-critical ego control rather than only route progress.

\subsection{TrajCond Form Analysis}
\label{sec:trajectory_representation}

\begin{figure*}[t]
    \centering
    \includegraphics[width=0.98\linewidth]{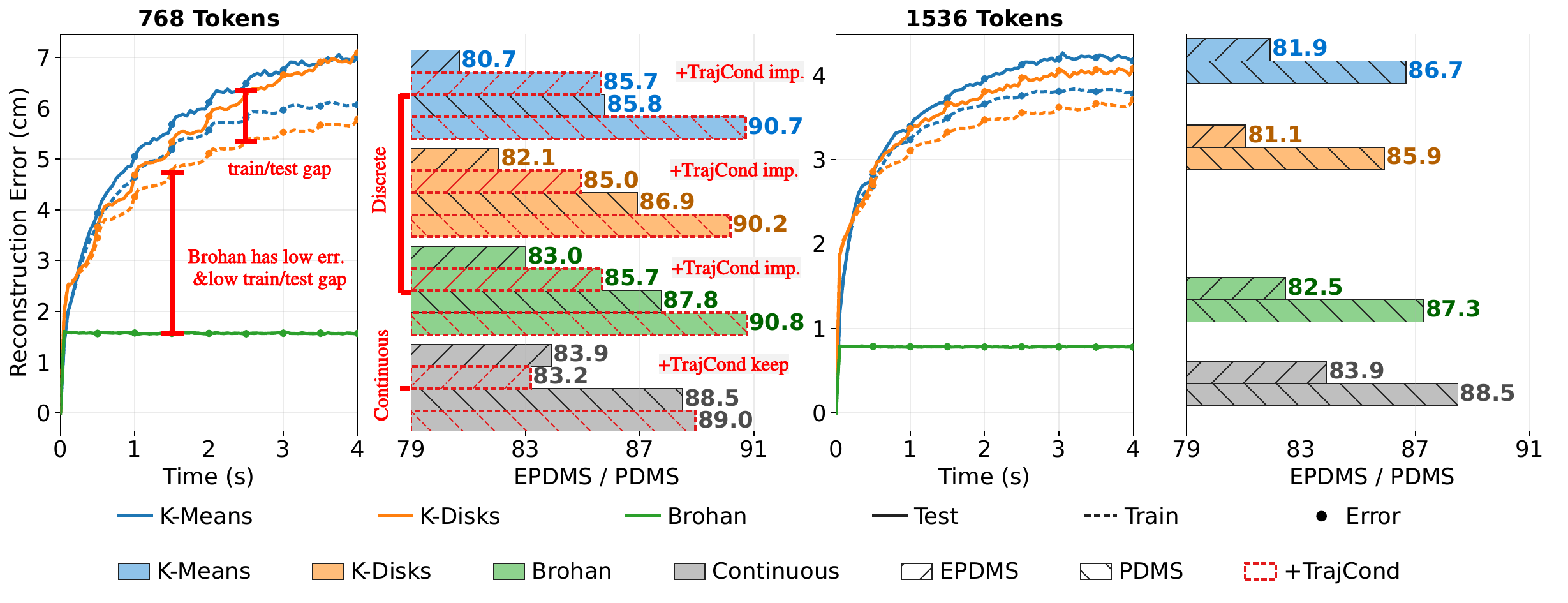}
    \caption{TrajCond form analysis for action tokens. Each codebook-size group pairs the reconstruction-error curve with the corresponding EPDMS/PDMS bars. The left axis reports trajectory reconstruction error; the right axis reports EPDMS/PDMS. Due to space limits, we move codebook-size=384 to  Supplement Fig.~\ref{fig:discrete-error-supp}.}
    \label{fig:discrete-error}
    \vspace{-20pt}
\end{figure*}

\noindent\textbf{Discretization error and generalization.}
Following Trajeglish~\cite{philiontrajeglish}, we measure the reconstruction error introduced by discretizing continuous trajectories into action tokens. Fig.~\ref{fig:discrete-error}/~\ref{fig:discrete-error-supp} (in supplement) shows a trend that $Q=384\rightarrow 768 \rightarrow 1536$, larger codebooks reduce reconstruction error and that 1D-Brohan consistently gives the smallest reconstruction error with a smaller train-test gap than clustering-based codebooks; Supplement Table~\ref{tab:js-distribution} quantifies the same shift.

\noindent\textbf{TrajCond as final prediction.}
When used as direct autoregressive output targets, all discrete representations remain weaker than continuous text waypoints (\texttt{grey bar} \textit{v.s.} \texttt{blue/yellow/green bars} in Fig.~\ref{fig:discrete-error}), even when Brohan gives the best discrete result . The K-Means/K-Disks comparison also shows that planning performance is not determined by oracle reconstruction error alone.

\noindent\textbf{Ablation on TrajCond form.}
We further evaluate the codebook-size 768 representations as trajectory-format semantic conditions. In Fig.~\ref{fig:discrete-error}, all three discrete tokenizers become much stronger when used as TrajCond, whereas using continuous trajectories as the condition gives only a limited gain. This confirms that action tokens are most useful as an intermediate interface rather than a replacement for final continuous prediction. Since 1D-Brohan gives both the lowest reconstruction error and the best TrajCond score, we use it as the default representation.

\subsection{TrajCond Semantic Analysis}
\label{sec:trajcond_semantic_validation}

\begin{table}[t]
\centering
\caption{TrajCond semantic validation on NAVSIM with Qwen3-VL-4B. Left: reasoning-dependence ablation. Except for the second row, which uses a standalone 6-epoch run, all left-table settings use the same 3+3 stage allocation. Right: perplexity analysis under pause-token replacement. In both tables, B, R, and C denote Brohan, Reason, and Continuous; Pause is written in full. Absolute perplexity values are provided in Supplement Table~\ref{tab:supp-perplexity}.}
\label{tab:cpr}
\newsavebox{\cprleftbox}
\newsavebox{\cprrightbox}
\newlength{\cprpanelheight}
\sbox{\cprleftbox}{%
\resizebox{0.63\linewidth}{!}{%
\begin{tabular}{l|c|c|c|c}
\toprule
\textbf{Method} & \textbf{Stage 1} & \textbf{Stage 2} & \textbf{RFT} & \textbf{EPDMS/PDMS} \\
\midrule
C-TC & R $\rightarrow$ C & C $\rightarrow$ R $\rightarrow$ C & -- & 83.2 / 89.0 \\
B-TC & R $\rightarrow$ B & B $\rightarrow$ R $\rightarrow$ C & -- & 84.9 / 90.6 \\
\midrule
B-TC w/o TrajCond & R $\rightarrow$ C (6 ep.) & -- & -- & 83.9 / 88.5 \\
B-TC w/o R & R $\rightarrow$ B & B $\rightarrow$ Pause $\rightarrow$ C & -- & 85.3 / 90.5 \\
\midrule
B-TC w/o R & R $\rightarrow$ B & B $\rightarrow$ Pause $\rightarrow$ C & \checkmark & 84.1 / 89.1 \\
B-TC & R $\rightarrow$ B & B $\rightarrow$ R $\rightarrow$ C & \checkmark & \textbf{86.8 / 92.0} \\
\bottomrule
\end{tabular}
}
}
\sbox{\cprrightbox}{%
\resizebox{0.35\linewidth}{!}{%
\begin{tabular}{l|l|l|c}
\toprule
\textbf{Setting} & \textbf{Cond.} & \textbf{Target} & \textbf{$\Delta$PPL} \\
\midrule
\multirow{1}{*}{nuScenes} & B$\rightarrow$Pause & C & +0.0094 \\
\midrule
\multirow{3}{*}{NAVSIM} & B$\rightarrow$Pause & C & +0.0011 \\
 & R$\rightarrow$Pause & C & +0.0012 \\
 & R$\rightarrow$Pause & B & +0.0002 \\
\bottomrule
\end{tabular}
}
}
\setlength{\cprpanelheight}{\dimexpr\ht\cprleftbox+\dp\cprleftbox\relax}
\begin{minipage}[c][\cprpanelheight][c]{0.63\linewidth}
\centering
\usebox{\cprleftbox}
\end{minipage}\hfill
\begin{minipage}[c][\cprpanelheight][c]{0.35\linewidth}
\centering
\usebox{\cprrightbox}
\end{minipage}
\end{table}

\noindent\textbf{Reasoning Dependence and Perplexity Analysis.} Pause tokens follow Goyal et al.~\cite{goyal2024think} and serve only as a structure-preserving placeholder, \textit{i.e.}, be used to replace the meaningful content with \texttt{[PAUSE]} proxy token. \textbf{(I) The left half of Table}~\ref{tab:cpr} tests whether TrajCond truly learns semantics rather than acting only as a geometric auxiliary signal. The Stage-1 comparison already supports this point: removing TrajCond and reverting to continuous regression degrades performance (\texttt{B-TC}>\texttt{B-TC w/o TrajCond}). We then keep the same two-stage SFT recipe but remove the Stage-2 textual reason and replace it with pause tokens, so the second-stage prediction must rely on TrajCond instead of re-reading explicit reasoning text. The resulting gain (\texttt{B-TC}<\texttt{B-TC w/o R}) suggests that once reasoning semantics have been absorbed into TrajCond, the trajectory-structured condition can provide cleaner guidance than injecting reason text again in Stage 2. \textbf{(II) The right half of }Table~\ref{tab:cpr} provides a complementary decoding view. On nuScenes, replacing action tokens with pause tokens increases the perplexity of the final continuous trajectory. On NAVSIM Stage 2, replacing either action tokens or reasoning with pause tokens also increases perplexity, with the reasoning replacement causing the largest degradation. On NAVSIM Stage 1, replacing reasoning with pause tokens similarly increases the perplexity of TrajCond prediction itself. Together with the reasoning-dependence ablation, these results suggest that both TrajCond and reasoning reduce the uncertainty of trajectory decoding, while TrajCond serves as a semantically meaningful trajectory-geometric condition rather than a purely geometric auxiliary signal. \textbf{(III)} \texttt{C-TC}<\texttt{B-TC} proves that the gains is not from longer output/auxiliary loss but from the semantic nature of Brohan action tokens. \textbf{(IV)} The RFT result further clarifies the role of TrajCond beyond the two SFT stages. Starting from the w/o-R variant, GRPO reduces performance, whereas the full B-TC + RFT pipeline remains strongest. This suggests that TrajCond is a strong complement to reasoning, but not a full replacement for textual reasoning during RFT, likely because RFT is harder to improve without reliable CoT reasoning support~\cite{rahman2026can}.

\begin{figure}[t]
    \centering
    \begin{minipage}[t]{0.49\linewidth}
        \centering
        \includegraphics[width=\linewidth]{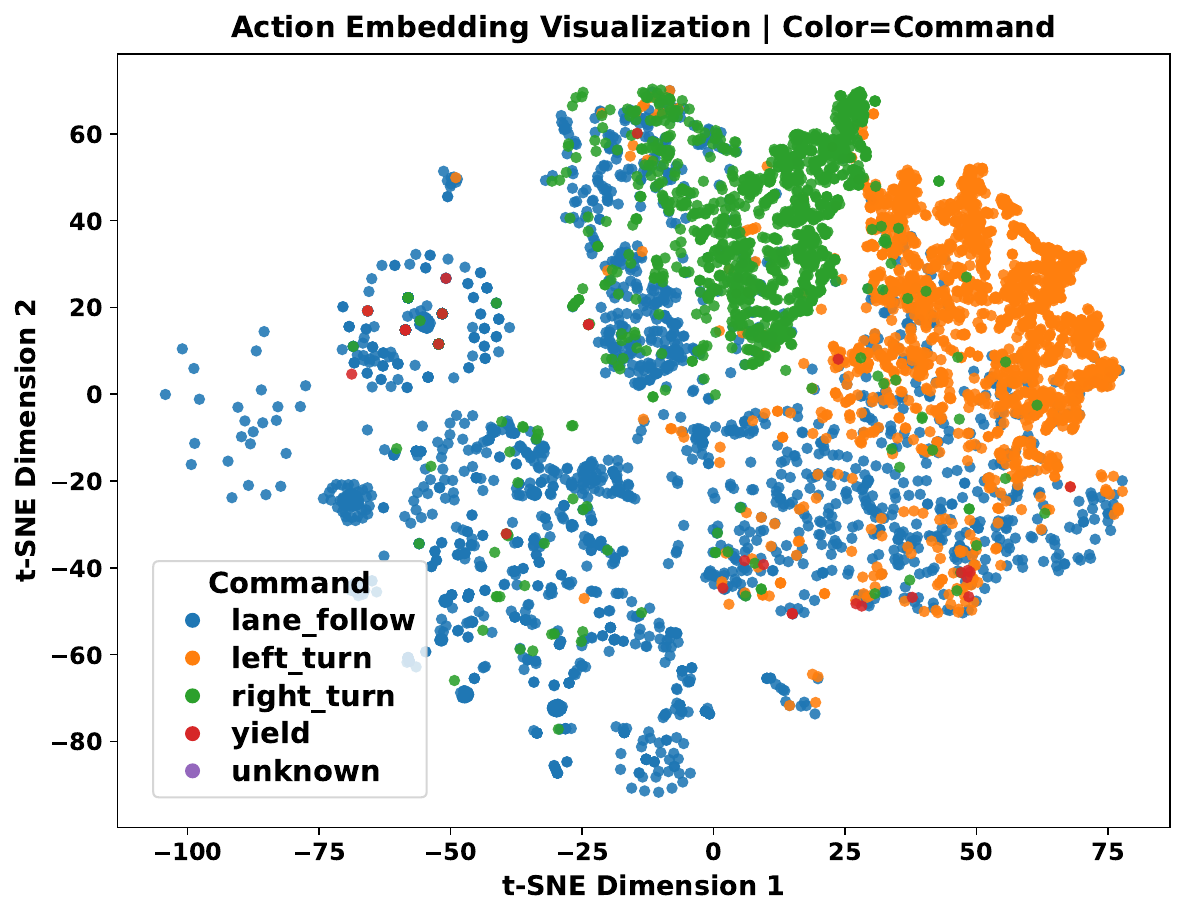}
    \end{minipage}\hfill
    \begin{minipage}[t]{0.49\linewidth}
        \centering
        \includegraphics[width=\linewidth]{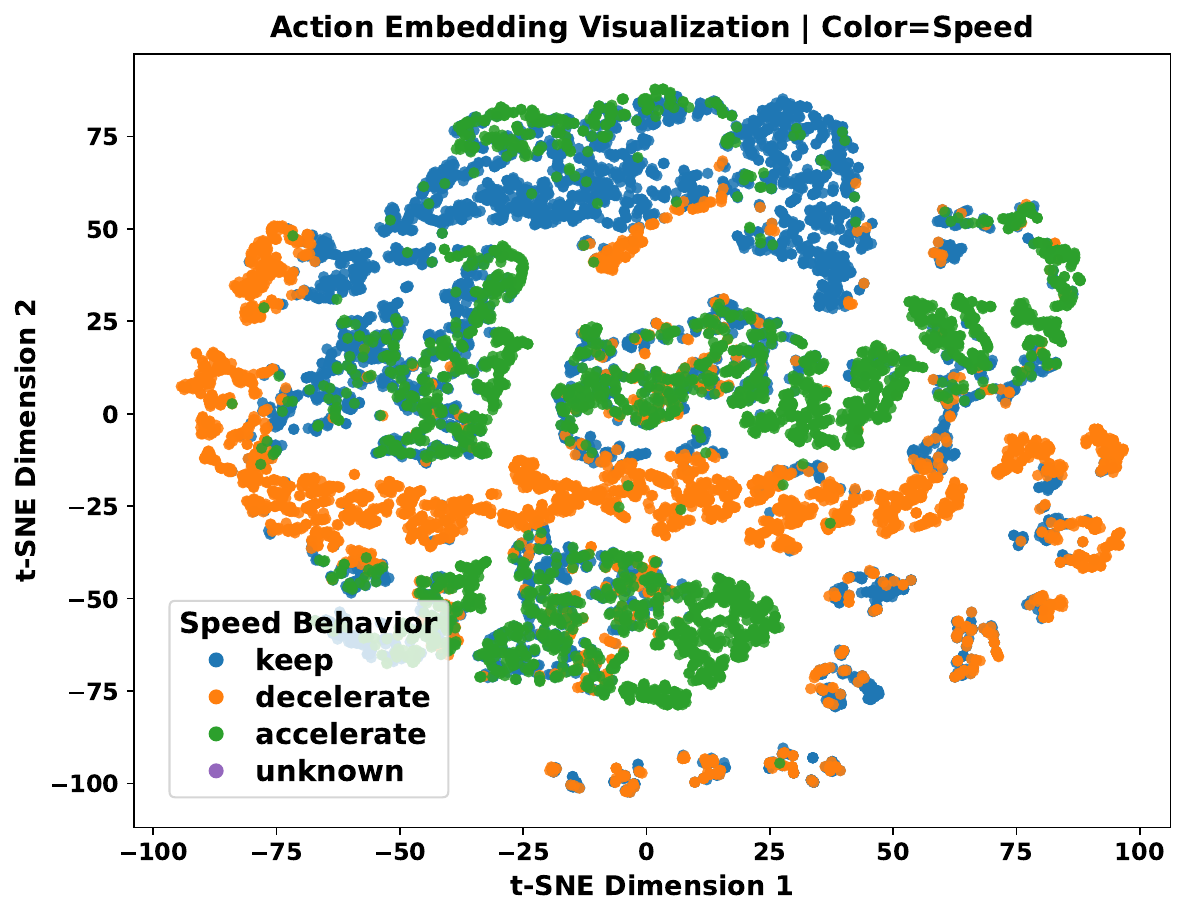}
    \end{minipage}
    \caption{t-SNE visualization of hidden states from the discrete trajectory used as TrajCond. Left: points are colored by instruction motion direction. Right: points are colored by speed behavior.}
    \label{fig:trajcond-hidden-state}
    \vspace{-15pt}
\end{figure}

\noindent\textbf{T-SNE Visualization.}
We further inspect the hidden states of the discrete trajectory used as TrajCond on NAVSIM navtest. We extract the motion direction/speed behavior from the CoT reasoning meta behavior. In Fig.~\ref{fig:trajcond-hidden-state}, the left panel shows grouping by motion direction, while the right panel shows grouping by speed behavior.  The t-SNE structure reveals tendency boundaries between different motion types. Interestingly, samples within the same action category still exhibit substantial spread rather than collapsing into a single compact center, suggesting that TrajCond captures rich semantic variation beyond mere action labels.

\section{Conclusion \& Limitation}
\label{sec:discussion}

\noindent \textbf{Conclusion.} In this paper, we introduce a discrete trajectory as TrajCond and propose a two-stage framework for using it in VLA planning. Our analyses lead to three conclusions. First, direct regression on discrete trajectories is weaker than direct regression on continuous trajectories. Second, when used as TrajCond, discrete trajectories are more effective than continuous-trajectory conditions and consistently improve final continuous trajectory prediction. Third, this gain comes from TrajCond absorbs useful planning semantics rather than auxiliary losses/ output length. 

\noindent \textbf{Limitation.} AutoVLA-D adds decoding/training cost through the intermediate trajectory condition,
  and some low-frequency action tokens may be insufficiently learned.

\section*{Acknowledge}
This work is funded by DiDi and Wuxi Research Institute of Applied
Technologies, Tsinghua University under Grant 20242001120.

\bibliographystyle{plainnat}
\bibliography{reference}

\newpage
\appendix

\section{Computation Resources}
All experiments—including two-stage model training, reinforcement fine-tuning, and comprehensive benchmark evaluations—were conducted on a high-performance internal computing cluster. Specifically, the compute node utilized is equipped with 8 NVIDIA H100 GPUs, 2TB of system memory (RAM), and 192 CPU cores. Under this hardware configuration, the complete suite of experiments required approximately one week of continuous execution time.

\section{Benchmark Details}
\label{sec:supp-benchmark-details}

We provide brief benchmark details for the three evaluation settings used in the main paper.

\noindent\textbf{NAVSIM navtest.}
Following Curious-VLA~\cite{chen2026devil}, we use NAVSIM navtest as the primary large-scale non-reactive closed-loop simulation benchmark. It focuses on planning evaluation under challenging driving scenarios and is reported here under the same protocol family as ReCogDrive~\cite{li2025recogdrive}, using PDMS and EPDMS on the official navtest split.

\noindent\textbf{nuScenes.}
Following Curious-VLA~\cite{chen2026devil} and Impromptu VLA~\cite{chi2025impromptu}, we use nuScenes as the real-world open-loop benchmark for auxiliary generalization evaluation. nuScenes contains 1000 driving scenes of 20 seconds each and is evaluated here under the standard 3-second planning metrics, including L2 distance error and collision rate.

\noindent\textbf{NeuroNCAP.}
Following ImagiDrive~\cite{li2025imagidrive}, we use NeuroNCAP as the safety-critical closed-loop benchmark. NeuroNCAP is a photorealistic simulator built on nuScenes and is designed to stress-test driving policies under interactive closed-loop rollout. We report NeuroNCAP Score together with collision rate under the same evaluation setting used in prior VLA comparison work.

\section{System Prompt}
\label{sec:supp-system-prompt}

For NAVSIM, we use Qwen3-VL-235B to generate pseudo-labeled CoT reasoning. The prompt is adapted from DriveTeach-VLA by keeping the reasoning subtasks and removing the future-trajectory-prediction task.

\begin{figure*}[ht]
\centering
\fbox{%
\begin{minipage}{0.94\linewidth}
\footnotesize
You are an expert driver. Suppose you are driving. Let's complete the following tasks step by step.

\textbf{Input:} front-view image, high-level intent $\in\{\text{GO STRAIGHT},\text{TURN RIGHT},\text{TURN LEFT}\}$, 1.5s past ego trajectory, and 4s expert future trajectory. Each trajectory point is $(x,y,\text{heading})$.

\textbf{Task 1: Critical objects.} For each class, answer yes/no on whether it can influence the ego future path: nearby vehicle, conflicting pedestrian, cyclist, construction, traffic element, weather condition, road hazard, emergency vehicle, animal, special vehicle, conflicting vehicle, door-opening vehicle.

\textbf{Task 2: Explanation.} Give a concise explanation of the expert 4s trajectory. Mention only yes-marked classes and explain how they affect the trajectory.

\textbf{Task 3: Meta-behavior.} Select one speed label from \{keep, accelerate, decelerate, other\} and one command label from \{straight, yield, left turn, right turn, lane follow, lane change left, lane change right, reverse, other\}.

\textbf{Output:} strict JSON with keys \texttt{critical\_objects}, \texttt{explanation}, and \texttt{meta\_behaviour}; no extra keys, markdown, or commentary.
\end{minipage}
}
\caption{Prompt for Qwen3-VL-235B to pseudo-label NAVSIM CoT reasoning data.}
\label{fig:sys-cot-label}
\end{figure*}

\noindent\textbf{NAVSIM Stage-1 user prompt.}
The Stage-1 NAVSIM training prompt asks the model to predict TrajCond action tokens from the image, intent, and past trajectory:
\begin{Verbatim}
You are an expert driver. Suppose you are driving. Let's complete the following tasks step by step.
Input:
- 1 frame of front-view image collected from the ego-vehicle at the present timestep
Picture 1: <image> the front view of the ego-vehicle
- Current high-level intent (string): go straight
- 1.5-second past trajectory(3 steps at 2 Hz):    - t-3: (0.0, 0.0, 0.0)    - t-2: (0.0, 0.0, 0.0)    - t-1: (0.0, 0.0, 0.0)    - t-0: (0.0, 0.0, 0.0)
Each trajectory point format: (x:float, y:float, heading:float)
Task 1: Critical Objects and Conditions Detection
Decide whether at least one critical instance of each class could influence the ego-vehicle's future path (no omissions). A vehicle can be a car, bus, truck, motorcyclist, scooter, etc. traffic_element includes traffic signs and traffic lights. road_hazard may include hazardous road conditions, road debris, obstacles, etc. A conflicting_vehicle is a vehicle that may potentially conflict with the ego's future path. Output "yes" or "no" for every class (no omissions).
   Object classes to audit:
     - nearby_vehicle
     - conflicting_pedestrian
     - cyclist
     - construction
     - traffic_element
     - weather_condition
     - road_hazard
     - emergency_vehicle
     - animal
     - special_vehicle
     - conflicting_vehicle
     - door_opening_vehicle

Task 2: Natural Language Explanation
(thinking output should be wrapped in <thinking>...</thinking>)
Compose a 100-word concise natural-language description of the optimal future 4-second trajectory for the ego vehicle that the expert driver (you) plans and explain why the expert driver plans to execute this trajectory.
   - Mention only the classes you marked "yes" in the previous task.
   - Describe how each of those critical objects or conditions influences the optimal trajectory.
   - Do not invent objects or conditions not present in the input.

Task 3: Meta-Behaviour Selection
Assign exactly one category from each list. Choose the label that best summarises the overall behaviour of the optimal future trajectory:
   - speed ? { keep, accelerate, decelerate }
   - command ? { straight, yield, left_turn, right_turn, lane_follow, lane_change_left, lane_change_right, reverse }
   Choose the label that best summarises the overall behaviour of the optimal future trajectory.
   - If none fits, use 'other', but do this sparingly.

Task 4: Future Trajectory Prediction
(answer output should be wrapped in <answer>...</answer>)
Given the input, critical objects/conditions, natural language explanation, and meta-behaviour, predict the optimal 4-second future trajectory (8 steps at 2 Hz) of the ego vehicle. Predict 8 groups of three future trajectory action tokens in [AT, ...] format.
Output format (strict JSON, no extra keys, no markdown codeblock chars(```), no commentary):
{
  "critical_objects": {
    "nearby_vehicle": "yes | no",
    "conflicting_pedestrian": "yes | no",
    "cyclist": "yes | no",
    "construction": "yes | no",
    "traffic_element": "yes | no",
    "weather_condition": "yes | no",
    "road_hazard": "yes | no",
    "emergency_vehicle": "yes | no",
    "animal": "yes | no",
    "special_vehicle": "yes | no",
    "conflicting_vehicle": "yes | no",
    "door_opening_vehicle": "yes | no"
  },
  "explanation": "100-word description that references only the classes marked 'yes'",
  "meta_behaviour": {
    "speed": "keep | accelerate | decelerate | other",
    "command": "straight | yield | left_turn | right_turn | lane_follow | lane_change_left | lane_change_right | reverse | other"
  },
  "future_trajectory": "<answer>[AT, ...]</answer>"
}
\end{Verbatim}

\noindent\textbf{NAVSIM Stage-2 user prompt.}
The Stage-2 NAVSIM training prompt asks the model to use the discrete action-token plan, environment reasoning, and driving logic before producing the final continuous trajectory:
\begin{Verbatim}
You are an expert driver. Suppose you are driving. Let's complete the following tasks step by step.
Input:
- 1 frame of front-view image collected from the ego-vehicle at the present timestep
Picture 1: <image> the front view of the ego-vehicle
- Current high-level intent (string): go straight
- 1.5-second past trajectory(3 steps at 2 Hz):    - t-3: (0.0, 0.0, 0.0)    - t-2: (0.0, 0.0, 0.0)    - t-1: (0.0, 0.0, 0.0)    - t-0: (0.0, 0.0, 0.0)
Each trajectory point format: (x:float, y:float, heading:float)
Task 1: Discrete Action Trajectory Planning
(answer output should be wrapped in <answer>...</answer>)
Based on the road geometry and intent, plan a naive "Discrete Action Trajectory" (8 steps at 2 Hz). Assume no dynamic obstacles (e.g., follow lane center, maintain speed).
Predict 8 future movement actions in [AT, ...] format.

Task 2: Critical Objects and Conditions Detection
Identify critical objects/conditions that might force a deviation from the Discrete Action Trajectory. Output "yes" or "no" for every class.
   Object classes to audit:
     - nearby_vehicle
     - conflicting_pedestrian
     - cyclist
     - construction
     - traffic_element
     - weather_condition
     - road_hazard
     - emergency_vehicle
     - animal
     - special_vehicle
     - conflicting_vehicle
     - door_opening_vehicle

Task 3: Meta-Behaviour Selection
Assign exactly one category from each list to describe the *Final* behavior:
   - speed ? { keep, accelerate, decelerate }
   - command ? { straight, yield, left_turn, right_turn, lane_follow, lane_change_left, lane_change_right, reverse }


Task 4: Natural Language Explanation
(thinking output should be wrapped in <thinking>...</thinking>)
Compose a 100-word concise description explaining the driving process.
   - Contrast the Discrete Action Plan with the environment.
   - Explain why the Discrete Action Plan must be adjusted to the Final Plan based on the visual evidence.

Task 5: Text Nuemeric Trajectory Planning
(answer output should be wrapped in <answer>...</answer>)
Based on the reasoning in Task 4, plan the optimal "Text Nuemeric Trajectory" (8 steps at 2 Hz) that accounts for all safety factors.
Predict 8 trajectory points in [PT, ...] format.

Output format (strict JSON, no extra keys, no markdown codeblock chars(```), no commentary):
{
  "discrete_action_trajectory": "<answer>[AT, ...]</answer>",
  "critical_objects": {
    "nearby_vehicle": "yes | no",
    "conflicting_pedestrian": "yes | no",
    "cyclist": "yes | no",
    "construction": "yes | no",
    "traffic_element": "yes | no",
    "weather_condition": "yes | no",
    "road_hazard": "yes | no",
    "emergency_vehicle": "yes | no",
    "animal": "yes | no",
    "special_vehicle": "yes | no",
    "conflicting_vehicle": "yes | no",
    "door_opening_vehicle": "yes | no"
  },
  "meta_behaviour": {
    "speed": "keep | accelerate | decelerate | other",
    "command": "straight | yield | left_turn | right_turn | lane_follow | lane_change_left | lane_change_right | reverse | other"
  },
  "explanation": "100-word description of the driving logic",
  "future_trajectory": "<answer>[PT, ...]</answer>"
}
\end{Verbatim}

\noindent\textbf{nuScenes and NeuroNCAP user prompt.}
For nuScenes and NeuroNCAP, Stage 1 and Stage 2 share the same user prompt template and differ only in the expected output target. The first line in the file is the prompt, followed by example Stage-1 and Stage-2 outputs:
\begin{Verbatim}
You are an autonomous driving agent. You have access to a front view camera image of a vehicle <image>. Your task is to do your best to predict future movement actions for the vehicle over the next 6 timesteps, given the vehicle's intent inferred from the images. Provided are the previous ego vehicle status recorded over the last 3.0 seconds (at 0.5-second intervals). This includes the x and y coordinates of the ego vehicle. Positive x means forward direction while positive y means leftwards. The data is presented in the format [x, y]:. (t-3.0s) [-19.78, -0.68], Acceleration: X +0.72, Y -0.43 m/s^2, Velocity: +5.72 m/s, Steering angle: -0.21 (positive: left turn, negative: right turn), (t-2.5s) [-16.86, -0.34], Acceleration: X +0.71, Y -1.03 m/s^2, Velocity: +5.93 m/s, Steering angle: -0.62 (positive: left turn, negative: right turn), (t-2.0s) [-13.73, -0.12], Acceleration: X +0.83, Y -1.07 m/s^2, Velocity: +6.19 m/s, Steering angle: -0.53 (positive: left turn, negative: right turn), (t-1.5s) [-10.50, -0.02], Acceleration: X +1.03, Y -0.69 m/s^2, Velocity: +6.52 m/s, Steering angle: -0.20 (positive: left turn, negative: right turn), (t-1.0s) [-7.05, +0.01], Acceleration: X +1.08, Y -0.41 m/s^2, Velocity: +6.87 m/s, Steering angle: -0.10 (positive: left turn, negative: right turn), (t-0.5s) [-3.42, 0.00], Acceleration: X +1.00, Y -0.38 m/s^2, Velocity: +7.26 m/s, Steering angle: -0.04 (positive: left turn, negative: right turn), (t-0.0s) [0.00, 0.00], Acceleration: X +0.95, Y -0.43 m/s^2, Velocity: +7.56 m/s, Steering angle: -0.02 (positive: left turn, negative: right turn)



------------stage1 output------------


<PLANNING>Predicted future movement details for the next 3.0 seconds (sampled at 0.5-second intervals), including movement actions in x, y directions and heading rotation. For each timestep, only output three action tokens in brackets separated by comma. The action output is formatted as [<action_x>,<action_y>,<action_h>]: [<action_110>,<action_382>,<action_636>], [<action_110>,<action_383>,<action_638>], [<action_116>,<action_387>,<action_637>], [<action_120>,<action_388>,<action_638>], [<action_131>,<action_384>,<action_637>], [<action_124>,<action_383>,<action_637>]</PLANNING>


------------stage2 output------------


<PLANNING>Predicted future movement details for the next 3.0 seconds (sampled at 0.5-second intervals), including movement actions in x, y directions and heading rotation. For each timestep, only output three action tokens in brackets separated by comma. The action output is formatted as [<action_x>,<action_y>,<action_h>]: [<action_110>,<action_382>,<action_636>], [<action_110>,<action_383>,<action_638>], [<action_116>,<action_387>,<action_637>], [<action_120>,<action_388>,<action_638>], [<action_131>,<action_384>,<action_637>], [<action_124>,<action_383>,<action_637>]
Now I'm going to accumulate the 6 action groups and convert them into text waypoints for each time step. Let's conduct comprehensive safety assessment, identify risks like collision, overturn, boundary crossing or leaving drivablle area, mark high-risk objects, and propose optimized trajectories to avoid dangers while retaining the original motion purpose. Predicted future movement details for the next 3 seconds (sampled at 0.5-second intervals), including BEV location in x and y directions (in meters). Larger x means forward direction while larger y means leftwards. The output is formatted as [x, y]: [+3.97, -0.03], [+8.00, -0.07], [+12.23, -0.08], [+16.65, -0.09], [+21.53, -0.10], [+26.11, -0.12]</PLANNING>
\end{Verbatim}

\section{Trajectory Differential Illustration}
\label{sec:supp-traj-differential}

\begin{figure}[H]
\centering
\includegraphics[width=0.82\linewidth]{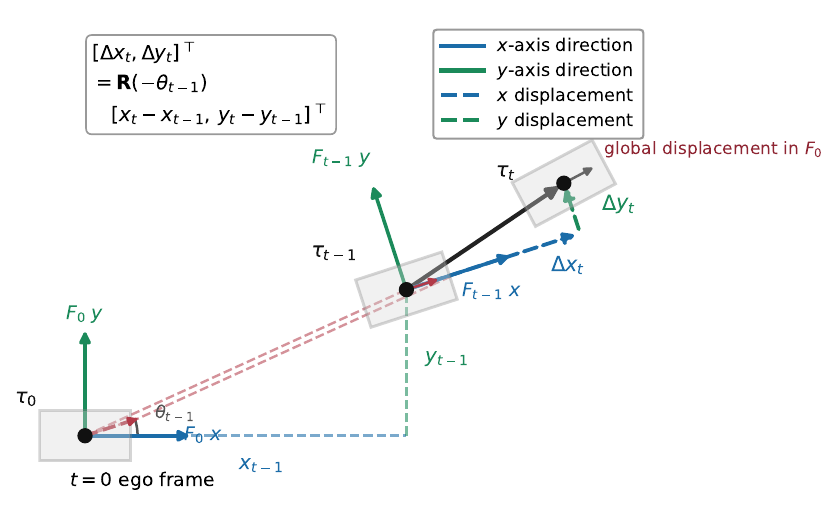}
\caption{Illustration of local-frame differential motion. The displacement between $\tau_{t-1}$ and $\tau_t$ is first observed in the $t=0$ ego frame, then rotated by $\mathbf{R}(-\theta_{t-1})$ into the local frame at step $t-1$ to obtain $(\Delta x_t,\Delta y_t)$.}
\label{fig:local-frame-differential}
\end{figure}

\begin{figure}[H]
\centering
\includegraphics[width=0.82\linewidth]{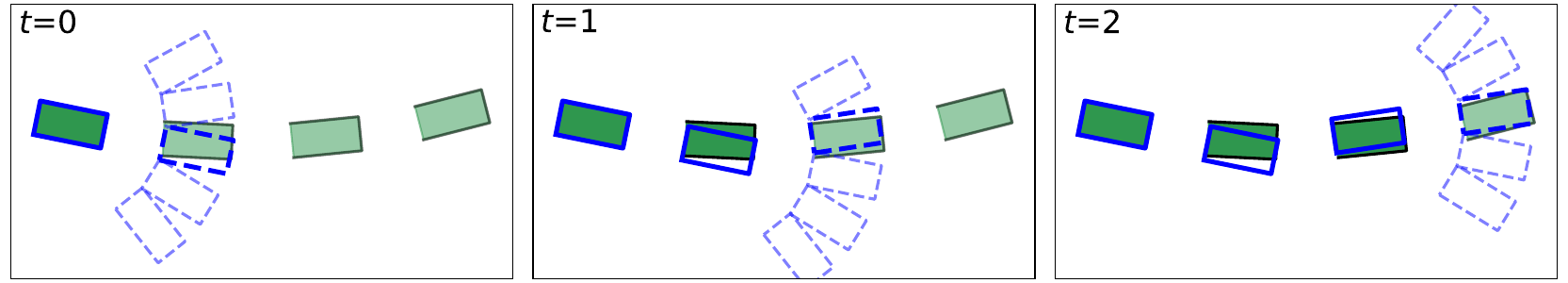}
\caption{Trajectory tokenization illustration \textbf{Copied From Trajeglish}~\cite{philiontrajeglish}. An example trajectory is shown in green. The raw representation of the tokenized trajectory is shown as boxes with blue outlines. States that have yet to be tokenized are light green. Token templates are optimized to minimize the error between the tokenized trajectories and the raw trajectories.}
\label{fig:trajeglish-tokenization}
\end{figure}

Fig.~\ref{fig:local-frame-differential} illustrates the coordinate transform used to compute differential motions. Although the trajectory waypoints are expressed in the ego frame at $t=0$, the transition from step $t-1$ to $t$ is represented in the local frame at step $t-1$. Fig.~\ref{fig:trajeglish-tokenization} visualizes the iterative trajectory-tokenization process, and Alg.~\ref{alg:brohan-token-assignment} gives the complete 1D-Brohan codebook construction and action-token assignment procedure used by AutoVLA-D.

\begin{algorithm}[H]
\caption{1D-Brohan codebook construction and action-token assignment}
\label{alg:brohan-token-assignment}
\begin{algorithmic}[1]
\REQUIRE Training trajectories $\mathcal{D}$, target trajectory $\tau=\{(x_t,y_t,\theta_t)\}_{t=1}^{T}$, and bin counts $\{Q_i\}_{i=1}^{3}$
\ENSURE Codebooks $\mathbf{x}$, $\mathbf{y}$, $\boldsymbol{\theta}$ and action-token sequence $\mathbf{a}=\{a_t\}_{t=1}^{T}$
\STATE Compute local-frame differential motions $\delta_t=(\Delta x_t,\Delta y_t,\Delta\theta_t)$ for all trajectories in $\mathcal{D}$ using Eq.~\ref{eq:local_frame_delta}:
\STATE $\Delta x_t=(x_t-x_{t-1})\cos\theta_{t-1}+(y_t-y_{t-1})\sin\theta_{t-1}$.
\STATE $\Delta y_t=-(x_t-x_{t-1})\sin\theta_{t-1}+(y_t-y_{t-1})\cos\theta_{t-1}$, $\Delta\theta_t=\theta_t-\theta_{t-1}$.
\STATE We adopt the 1D-Brohan codebook in Alg.~\ref{alg:brohan-token-assignment}. For each DoF, take the 1st and 99th quantiles of the training differentials as the discretization range.
\STATE $r_i=\bigl[\operatorname{Quantile}_{1\%}(\{\delta_t^i\}),\operatorname{Quantile}_{99\%}(\{\delta_t^i\})\bigr]$, $i\in\{1,2,3\}$.
\STATE Uniformly divide the three ranges into $Q_1,Q_2,Q_3$ bins to obtain codebooks $\mathbf{x}$, $\mathbf{y}$, and $\boldsymbol{\theta}$.
\STATE Initialize $\hat{\tau}_0=(0,0,0)$.
\FOR{$t=1$ to $T$}
\STATE Select the four nearest tokens in each DoF by absolute residual from the local-frame differential motion.
\STATE Enumerate the resulting candidate set $\mathcal{K}_t$ of $4\times4\times4$ tuples.
\FOR{each $\mathbf{k}=(k_1,k_2,k_3)\in\mathcal{K}_t$}
\STATE Render $\bar{\tau}_t(\mathbf{k})=(\bar{x}_t(\mathbf{k}),\bar{y}_t(\mathbf{k}),\bar{\theta}_t(\mathbf{k}))$ from $\hat{\tau}_{t-1}$ using Eq.~\ref{eq:action_recon}:
\STATE $\left[\begin{smallmatrix}\bar{x}_t(\mathbf{k})\\ \bar{y}_t(\mathbf{k})\end{smallmatrix}\right]=\left[\begin{smallmatrix}\hat{x}_{t-1}\\ \hat{y}_{t-1}\end{smallmatrix}\right]+\mathbf{R}(\hat{\theta}_{t-1})\left[\begin{smallmatrix}x_{k_1}\\ y_{k_2}\end{smallmatrix}\right]$.
\STATE $\bar{\theta}_t(\mathbf{k})=\hat{\theta}_{t-1}+\theta_{k_3}$.
\ENDFOR
\STATE $\mathbf{k}_t^\ast \leftarrow \arg\min_{\mathbf{k}\in\mathcal{K}_t} d_{\mathrm{corner}}(\bar{\tau}_t(\mathbf{k}),\tau_t)$.
\STATE $a_t\leftarrow\mathbf{k}_t^\ast$ and $\hat{\tau}_t\leftarrow\bar{\tau}_t(\mathbf{k}_t^\ast)$.
\ENDFOR
\RETURN $\mathbf{a}$
\end{algorithmic}
\end{algorithm}

\begin{table}[H]
\centering
\scriptsize
\caption{Additional baselines on NAVSIM navtest under PDMS. These rows are moved from the main table to keep the main comparison focused on stronger VLA methods.}
\label{tab:supp-navsim-pdms-e2e}
\resizebox{0.82\linewidth}{!}{%
\begin{tabular}{llcccccc}
\toprule
Method & Base Model & NC$\uparrow$ & DAC$\uparrow$ & EP$\uparrow$ & TTC$\uparrow$ & C$\uparrow$ & PDMS$\uparrow$ \\
\midrule
Ego-MLP & -- & 93.1 & 78.3 & 63.2 & 84.0 & 100.0 & 66.4 \\
UniAD & -- & 97.7 & 92.8 & 79.2 & 92.8 & 100.0 & 84.0 \\
DiffusionDrive & -- & 98.2 & 96.2 & 82.2 & 94.7 & 100.0 & 88.1 \\
WoTE & -- & 98.5 & 96.8 & 81.9 & 94.4 & 99.9 & 88.3 \\
Qwen2.5-VL & Qwen2.5-VL-7B & 97.8 & 92.1 & 78.3 & 92.8 & 100.0 & 83.3 \\
InternVL3 & InternVL3-8B & 97.0 & 92.4 & 78.9 & 91.8 & 100.0 & 83.3 \\
\bottomrule
\end{tabular}
}
\end{table}

\begin{table}[H]
\centering
\scriptsize
\caption{Additional E2E baselines on NAVSIM navtest under EPDMS. These rows are moved from the main table to keep the main comparison focused on VLA methods.}
\label{tab:supp-navsim-epdms-e2e}
\resizebox{\linewidth}{!}{%
\begin{tabular}{lcccccccccc}
\toprule
Method & NC$\uparrow$ & DAC$\uparrow$ & DDC$\uparrow$ & TLC$\uparrow$ & EP$\uparrow$ & TTC$\uparrow$ & LK$\uparrow$ & C$\uparrow$ & EC$\uparrow$ & EPDMS$\uparrow$ \\
\midrule
Ego-MLP & 93.1 & 77.9 & 92.7 & 99.6 & 86.0 & 91.5 & 89.4 & 98.3 & 85.4 & 64.0 \\
VADv2 & 97.3 & 91.7 & 98.2 & 99.9 & 77.6 & 92.7 & 66.0 & 100.0 & 97.4 & 76.6 \\
TransFuser & 96.9 & 89.9 & 97.8 & 99.7 & 87.1 & 95.4 & 92.7 & 98.3 & 87.2 & 76.7 \\
HydraMDP++ & 97.2 & 97.5 & 99.4 & 99.6 & 83.1 & 96.5 & 94.4 & 98.2 & 70.9 & 81.4 \\
DriveSuprim & 97.5 & 96.5 & 99.4 & 99.6 & 88.4 & 96.6 & 95.5 & 98.3 & 77.0 & 83.1 \\
ARTEMIS & 98.3 & 95.1 & 98.6 & 99.8 & 81.5 & 97.4 & 96.5 & 98.3 & -- & 83.1 \\
\bottomrule
\end{tabular}
}
\end{table}

\section{Progressive Analysis Data}
\label{sec:supp-progressive-data}

Table~\ref{tab:supp-progressive-data} gives the numeric values plotted in the right panel of Table~\ref{tab:progressive}. The stage descriptions follow the same abbreviations as the main paper: Reason, Continuous, Brohan, and epochs are denoted by R, C, B, and ep., respectively.

\begin{table}[H]
\centering
\caption{Detailed progressive-analysis data behind Table~\ref{tab:progressive}.}
\label{tab:supp-progressive-data}
\resizebox{\linewidth}{!}{%
\begin{tabular}{l|l|l|c|cc}
\toprule
\textbf{Method} & \textbf{Stage 1} & \textbf{Stage 2} & \textbf{RFT} & \textbf{EPDMS$\uparrow$} & \textbf{PDMS$\uparrow$} \\
\midrule
Brohan (B) & R $\rightarrow$ B (6 ep.) & -- & -- & 82.4 & 86.8 \\
Continuous (C) & R $\rightarrow$ C (6 ep.) & -- & -- & 83.9 & 88.5 \\
Brohan TrajCond (1) & -- & B $\rightarrow$ R $\rightarrow$ C (6 ep.) & -- & 84.0 & 89.2 \\
Brohan TrajCond & R $\rightarrow$ B (3 ep.) & B $\rightarrow$ R $\rightarrow$ C (3 ep.) & -- & 85.6 & 90.8 \\
Continuous + RFT & R $\rightarrow$ C (6 ep.) & -- & \checkmark & 84.9 & 89.7 \\
Brohan TrajCond + RFT & R $\rightarrow$ B (3 ep.) & B $\rightarrow$ R $\rightarrow$ C (3 ep.) & \checkmark & 86.8 & 92.0 \\
\bottomrule
\end{tabular}
}
\end{table}

\section{Complete Trajectory-Representation Analysis}
\label{sec:supp-discrete-error}

\begin{figure}
    \centering
    \includegraphics[width=1\linewidth]{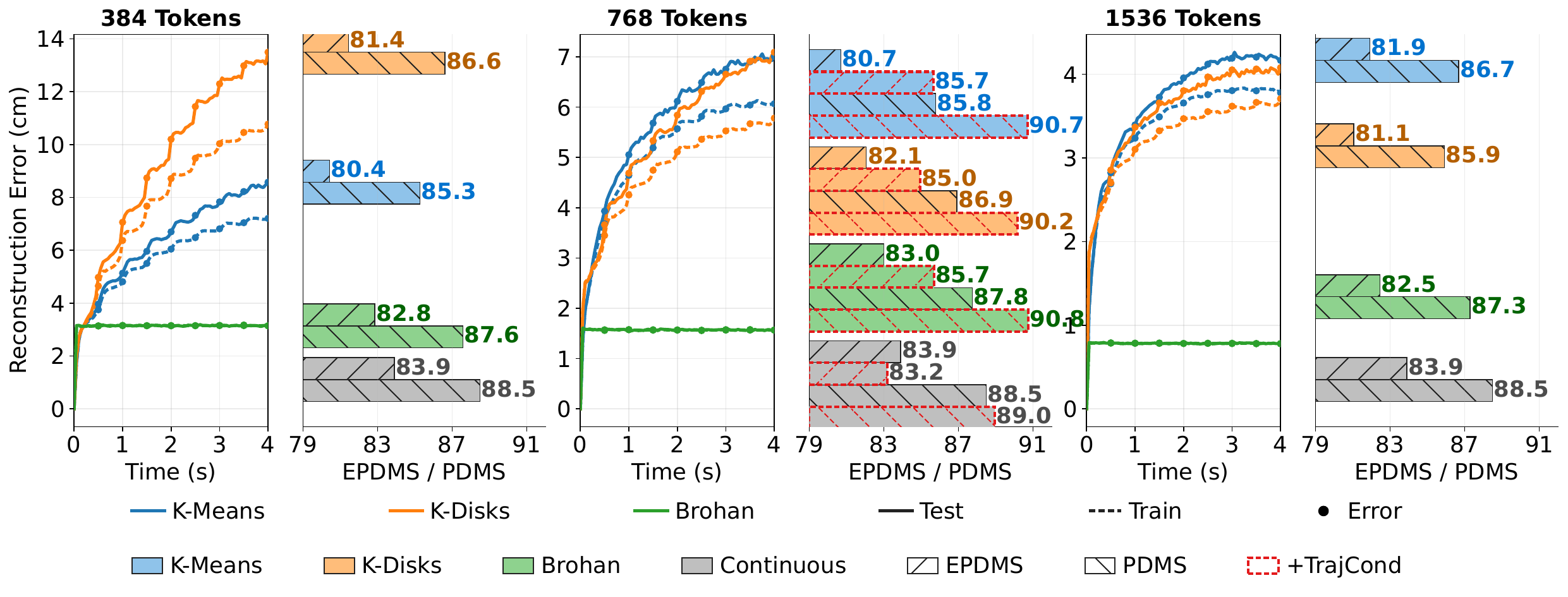}
    \caption{Complete trajectory-representation analysis across codebook sizes 384, 768, and 1536. The figure reports reconstruction error for each discretization method together with the corresponding PDMS/EPDMS performance bars.}
    \label{fig:discrete-error-supp}
\end{figure}

Fig.~\ref{fig:discrete-error-supp} extends the main-paper trajectory-representation analysis in Fig.~\ref{fig:discrete-error} by adding the smaller codebook-size 384 setting. The reconstruction-error curves show the same qualitative behavior as the main figure: larger codebooks reduce discretization error, 1D-Brohan remains the most stable discrete representation, and clustering-based K-Means and K-Disks show larger train-test gaps than rule-based Brohan. Continuous text waypoints are plotted as the zero-discretization-error reference because they do not require tokenization before trajectory regression.

Table~\ref{tab:supp-discrete-error-values} reports the reconstruction-error values behind the curves in Fig.~\ref{fig:discrete-error-supp}. Values are sampled at 1, 2, 3, and 4 seconds from the 80-step evaluation trajectory and are reported in centimeters.

\begin{table}[H]
\centering
\caption{Trajectory reconstruction error values behind Fig.~\ref{fig:discrete-error-supp}. Lower is better.}
\label{tab:supp-discrete-error-values}
\resizebox{\linewidth}{!}{%
\begin{tabular}{c|l|cccc|cccc}
\toprule
\multirow{2}{*}{\textbf{Codebook}} & \multirow{2}{*}{\textbf{Tokenizer}} &
\multicolumn{4}{c|}{\textbf{Train error (cm)}} &
\multicolumn{4}{c}{\textbf{Test error (cm)}} \\
\cmidrule(lr){3-6}\cmidrule(lr){7-10}
& & \textbf{1s} & \textbf{2s} & \textbf{3s} & \textbf{4s} & \textbf{1s} & \textbf{2s} & \textbf{3s} & \textbf{4s} \\
\midrule
384 & 3D K-Means & 4.81 & 6.04 & 6.81 & 7.21 & 5.13 & 6.70 & 7.83 & 8.58 \\
384 & 3D K-Disks & 6.37 & 8.71 & 10.03 & 10.77 & 7.06 & 10.20 & 12.29 & 13.49 \\
384 & 1D-Brohan & 3.16 & 3.15 & 3.15 & 3.14 & 3.15 & 3.14 & 3.15 & 3.14 \\
\midrule
768 & 3D K-Means & 4.64 & 5.57 & 5.97 & 6.07 & 5.05 & 6.12 & 6.75 & 7.01 \\
768 & 3D K-Disks & 4.25 & 5.11 & 5.53 & 5.78 & 4.69 & 5.84 & 6.65 & 7.09 \\
768 & 1D-Brohan & 1.57 & 1.56 & 1.57 & 1.57 & 1.58 & 1.57 & 1.57 & 1.56 \\
\midrule
1536 & 3D K-Means & 3.24 & 3.66 & 3.81 & 3.78 & 3.39 & 3.96 & 4.19 & 4.17 \\
1536 & 3D K-Disks & 3.10 & 3.47 & 3.62 & 3.71 & 3.37 & 3.80 & 4.05 & 4.08 \\
1536 & 1D-Brohan & 0.78 & 0.78 & 0.78 & 0.78 & 0.79 & 0.78 & 0.78 & 0.78 \\
\bottomrule
\end{tabular}
}
\end{table}

To quantify the train/test distribution shift behind this trend, Table~\ref{tab:js-distribution} reports the Jensen-Shannon (JS) divergence between the training and test action-token distributions for each tokenizer and codebook size. Lower JS divergence indicates a closer match between training and held-out token distributions. Across all codebook sizes, 1D-Brohan has the smallest divergence, while 3D K-Means and 3D K-Disks show larger train/test distribution shifts. This supports the main-paper interpretation that Brohan-style per-dimension discretization produces more stable action-token labels under held-out trajectories.

\begin{table}[t]
\centering
\caption{Train/test action-token distribution shift measured by Jensen-Shannon divergence. Lower is better.}
\label{tab:js-distribution}
\resizebox{0.75\linewidth}{!}{%
\begin{tabular}{c|ccc}
\toprule
\textbf{Codebook size} & \textbf{3D K-Means} & \textbf{3D K-Disks} & \textbf{1D-Brohan} \\
\midrule
384  & 0.0108958 & 0.0107432 & \textbf{0.0039045} \\
768  & 0.0126990 & 0.0135494 & \textbf{0.0044001} \\
1536 & 0.0167875 & 0.0175003 & \textbf{0.0051211} \\
\bottomrule
\end{tabular}
}
\end{table}

Table~\ref{tab:supp-discrete-score-values} reports the PDMS/EPDMS bar values shown in Fig.~\ref{fig:discrete-error-supp}. Direct output means that the action-token representation is used as the final autoregressive output target. TrajCond means that the representation is used as an intermediate condition before predicting the final continuous trajectory.

\begin{table}[H]
\centering
\caption{PDMS/EPDMS score values behind Fig.~\ref{fig:discrete-error-supp}. Higher is better.}
\label{tab:supp-discrete-score-values}
\resizebox{0.92\linewidth}{!}{%
\begin{tabular}{c|l|cc|cc}
\toprule
\multirow{2}{*}{\textbf{Codebook}} & \multirow{2}{*}{\textbf{Representation}} &
\multicolumn{2}{c|}{\textbf{Direct output}} &
\multicolumn{2}{c}{\textbf{TrajCond}} \\
\cmidrule(lr){3-4}\cmidrule(lr){5-6}
& & \textbf{EPDMS$\uparrow$} & \textbf{PDMS$\uparrow$} & \textbf{EPDMS$\uparrow$} & \textbf{PDMS$\uparrow$} \\
\midrule
384 & 3D K-Means & 80.43 & 85.27 & -- & -- \\
384 & 3D K-Disks & 81.44 & 86.61 & -- & -- \\
384 & 1D-Brohan & 82.85 & 87.58 & -- & -- \\
384 & Continuous & 83.90 & 88.49 & -- & -- \\
\midrule
768 & 3D K-Means & 80.69 & 85.78 & 85.65 & 90.72 \\
768 & 3D K-Disks & 82.06 & 86.92 & 84.95 & 90.17 \\
768 & 1D-Brohan & 83.00 & 87.75 & 85.69 & 90.75 \\
768 & Continuous & 83.90 & 88.49 & 83.19 & 88.96 \\
\midrule
1536 & 3D K-Means & 81.92 & 86.68 & -- & -- \\
1536 & 3D K-Disks & 81.06 & 85.92 & -- & -- \\
1536 & 1D-Brohan & 82.46 & 87.29 & -- & -- \\
1536 & Continuous & 83.90 & 88.49 & -- & -- \\
\bottomrule
\end{tabular}
}
\end{table}

The 384-codebook performance bars further explain why the main paper focuses on codebook sizes 768 and 1536. When action tokens are used as direct output targets, 3D K-Means reaches 80.43/85.27 EPDMS/PDMS, 3D K-Disks reaches 81.44/86.61, and 1D-Brohan reaches 82.85/87.58. These results follow the same ordering as the main figure: Brohan is the best discrete direct-output representation, but all discrete direct-output variants remain below the continuous text-waypoint baseline of 83.90/88.49. This supports the main-paper conclusion that direct autoregressive generation of discrete action tokens is limited by both discretization error and token-prediction difficulty.

Together with the 768 and 1536 settings in the main figure, the complete study supports the default design choice used by AutoVLA-D. The best role for action tokens is not to serve as the final output format; instead, they are most useful as an intermediate trajectory-geometric condition that guides the second-stage continuous planner.

\section{Complete Perplexity Values}
\label{sec:supp-perplexity}

Table~\ref{tab:supp-perplexity} reports the absolute perplexity values behind the main-paper pause-token replacement analysis in the right half of Table~\ref{tab:cpr}. All evaluations use temperature $= 1\mathrm{e}{-3}$ and top-$p = 1$. Following the main paper, we abbreviate Brohan, Reason, and Continuous as B, R, and C, while Pause is written in full.

\begin{table}[ht]
\centering
\caption{Absolute perplexity values for the pause-token replacement analysis. Lower is better.}
\label{tab:supp-perplexity}
\resizebox{0.82\linewidth}{!}{%
\begin{tabular}{l|l|l|c}
\toprule
\textbf{Setting} & \textbf{Condition} & \textbf{Target} & \textbf{Mean PPL} \\
\midrule
nuScenes & original & C & 1.0250 \\
nuScenes & B$\rightarrow$Pause & C & 1.0344 \\
\midrule
NAVSIM & original & C & 1.0057 \\
NAVSIM & B$\rightarrow$Pause & C & 1.0068 \\
NAVSIM & R$\rightarrow$Pause & C & 1.0069 \\
\midrule
NAVSIM & original & B & 1.0339 \\
NAVSIM & R$\rightarrow$Pause & B & 1.0341 \\
\bottomrule
\end{tabular}
}
\end{table}

\section{Intent-Grouped NC/DAC Statistics}
\label{sec:supp-nc-dac-stat}

Table~\ref{tab:supp-nc-dac-stat} provides the full statistics behind Fig.~\ref{fig:nc-dac-product-reduction}. We compare AutoVLA-D with the Continuous (C) baseline and report the number of samples where the NC-related collision term is zero, the DAC-related drivable-area term is zero, and their $NC{\times}DAC$ product is zero. Product rates use 12,146 samples as the denominator. Lower counts and rates indicate fewer samples whose multiplicative PDMS terms are zeroed.

\begin{table}[t]
\centering
\caption{Intent-grouped NC/DAC zero-count comparison against the Continuous (C) baseline. Product rates use 12,146 samples as the denominator, and reduction is computed as $(\mathrm{Baseline}-\mathrm{Ours})/\mathrm{Baseline}$.}
\label{tab:supp-nc-dac-stat}
\resizebox{\linewidth}{!}{%
\begin{tabular}{l|ccc|ccc|ccccc}
\toprule
\multirow{2}{*}{\textbf{Intent}} &
\multicolumn{3}{c|}{\textbf{NC zero count}} &
\multicolumn{3}{c|}{\textbf{DAC zero count}} &
\multicolumn{5}{c}{\textbf{$NC{\times}DAC$ zero}} \\
\cmidrule(lr){2-4}\cmidrule(lr){5-7}\cmidrule(lr){8-12}
& \textbf{Ours} & \textbf{C} & \textbf{Reduction} & \textbf{Ours} & \textbf{C} & \textbf{Reduction} & \textbf{Ours} & \textbf{C} & \textbf{Ours rate} & \textbf{C rate} & \textbf{Reduction} \\
\midrule
Go straight & 160 & 263 & 39.16\% & 153 & 209 & 26.79\% & 308 & 464 & 2.54\% & 3.82\% & 33.62\% \\
Turn left   & 29  & 52  & 44.23\% & 151 & 200 & 24.50\% & 178 & 246 & 1.47\% & 2.03\% & 27.64\% \\
Turn right  & 35  & 53  & 33.96\% & 85  & 162 & 47.53\% & 118 & 204 & 0.97\% & 1.68\% & 42.16\% \\
\bottomrule
\end{tabular}
}
\end{table}

\begin{figure}[t]
    \centering
    \includegraphics[width=1\linewidth]{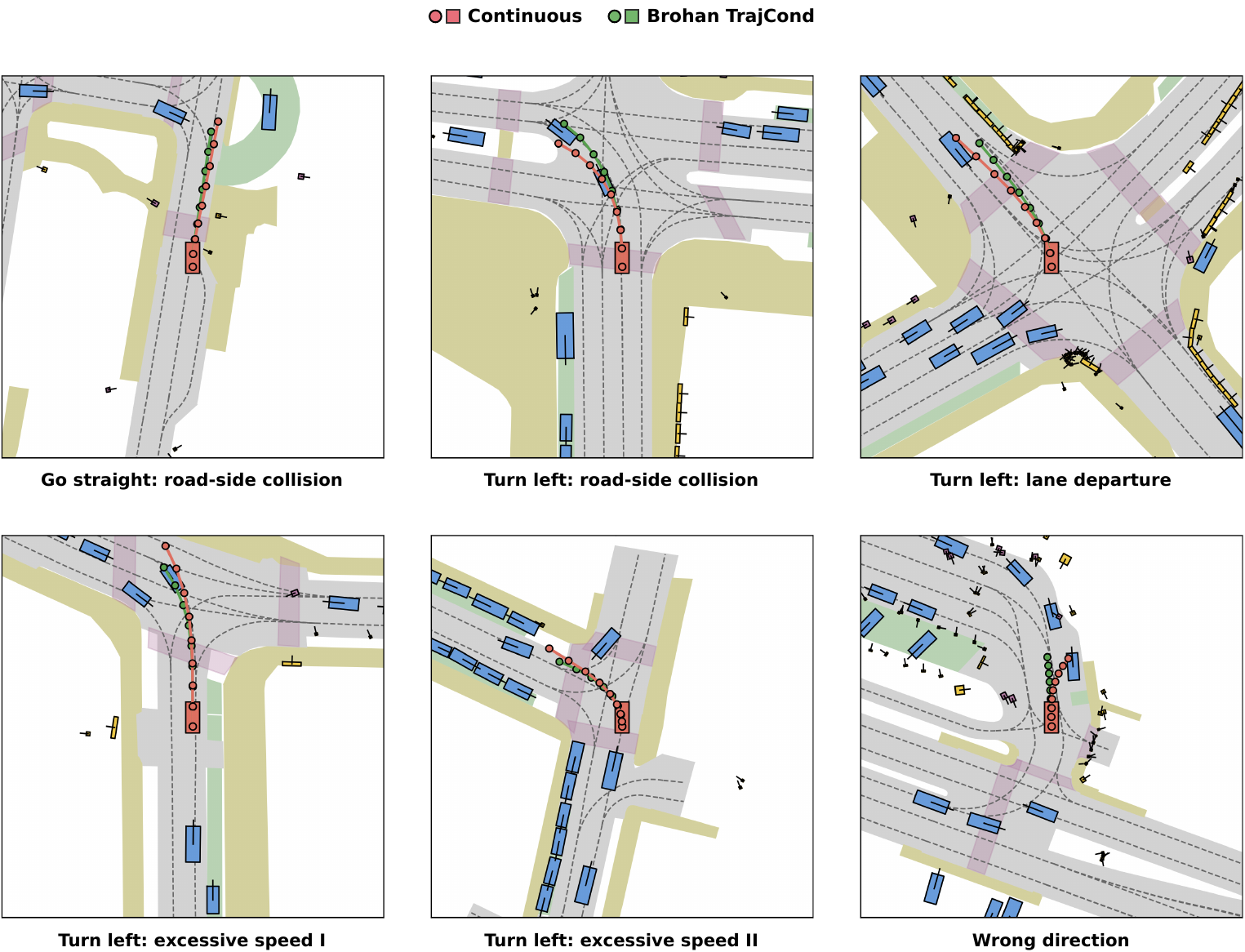}
    \caption{Additional collision-related examples for the NC/DAC analysis. The legend maps pink circular markers and square bars to Continuous, and green circular markers and square bars to Brohan TrajCond. The six panels preserve the original sample-image resolution after white-border cropping and show representative failure modes indicated by their panel labels: road-side collision, lane departure, excessive speed, and wrong direction.}
    \label{fig:nc-dac-supp-collision-examples}
\end{figure}

\clearpage
\newpage
\section*{NeurIPS Paper Checklist}

\begin{enumerate}

\item {\bf Claims}
    \item[] Question: Do the main claims made in the abstract and introduction accurately reflect the paper's contributions and scope?
    \item[] Answer: \answerYes{} 
    \item[] Justification: The abstract and introduction accurately outline the core contributions, including the AutoVLA-D framework, the 1D-Brohan trajectory representation, and the performance improvements on NAVSIM, nuScenes, and NeuroNCAP, which are all supported in Section 4.

\item {\bf Limitations}
    \item[] Question: Does the paper discuss the limitations of the work performed by the authors?
    \item[] Answer: \answerYes{} 
    \item[] Justification: Section 5 (Conclusion \& Limitation) explicitly discusses the limitations, noting that TrajCond should not be seen as a full replacement for explicit textual reasoning, especially during reinforcement fine-tuning.

\item {\bf Theory assumptions and proofs}
    \item[] Question: For each theoretical result, does the paper provide the full set of assumptions and a complete (and correct) proof?
    \item[] Answer: \answerNA{} 
    \item[] Justification: The paper does not include theoretical results or mathematical proofs.

    \item {\bf Experimental result reproducibility}
    \item[] Question: Does the paper fully disclose all the information needed to reproduce the main experimental results of the paper to the extent that it affects the main claims and/or conclusions of the paper (regardless of whether the code and data are provided or not)?
    \item[] Answer: \answerYes{} 
    \item[] Justification: Section 4.1 specifies the datasets, optimizer, learning rates, epochs, and hyperparameters, while Appendix C provides the exact system and user prompts required to reproduce the pseudo-labeled training data and model behavior.

\item {\bf Open access to data and code}
    \item[] Question: Does the paper provide open access to the data and code, with sufficient instructions to faithfully reproduce the main experimental results, as described in supplemental material?
    \item[] Answer: \answerNo{} 
    \item[] Justification: The research is based on publicly accessible datasets (NAVSIM, nuScenes, NeuroNCAP) and open-weights models (Qwen-VL). We are very sorry that we are currently unable to organize and make the code public, though we have provided sufficient information for easy reproduction.

\item {\bf Experimental setting/details}
    \item[] Question: Does the paper specify all the training and test details (e.g., data splits, hyperparameters, how they were chosen, type of optimizer) necessary to understand the results?
    \item[] Answer: \answerYes{} 
    \item[] Justification: Detailed training setups, including the use of AdamW, weight decay, warmup ratio, batch sizes, learning rates, and dataset splits, are provided in Section 4.1 and Appendix B.

\item {\bf Experiment statistical significance}
    \item[] Question: Does the paper report error bars suitably and correctly defined or other appropriate information about the statistical significance of the experiments?
    \item[] Answer: \answerNo{} 
    \item[] Justification: For NAVSIM, We only report single-run results without error bars due to insufficient computational respources, as training and evaluating large multimodal VLA models is highly computationally expensive, especially during the GRPO rollout stage. For nuScenes and NeuroNCAP, the reported results are the mean of 10 experimental runs with different random seeds, which mitigates randomness and statistical fluctuations.

\item {\bf Experiments compute resources}
    \item[] Question: For each experiment, does the paper provide sufficient information on the computer resources (type of compute workers, memory, time of execution) needed to reproduce the experiments?
    \item[] Answer: \answerYes{} 
    \item[] Justification: The Appendix A provides full details regarding the computational resources used, specifying the exact hardware configuration (an internal cluster with 8 NVIDIA H100 GPUs, 2TB RAM, 192 CPUs) and the approximate total execution time of one week.
    
\item {\bf Code of ethics}
    \item[] Question: Does the research conducted in the paper conform, in every respect, with the NeurIPS Code of Ethics \url{https://neurips.cc/public/EthicsGuidelines}?
    \item[] Answer: \answerYes{} 
    \item[] Justification: The research relies on standard, publicly available autonomous driving benchmarks and models, adhering to ethical norms and the NeurIPS Code of Ethics.

\item {\bf Broader impacts}
    \item[] Question: Does the paper discuss both potential positive societal impacts and negative societal impacts of the work performed?
    \item[] Answer: \answerNo{} 
    \item[] Justification: The paper focuses on the technical development of an autonomous driving VLA planner and does not dedicate a specific section to discussing broader societal impacts. Because our proposed method demonstrably reduces driving collisions and improves safety-critical control, its application inherently yields positive societal impacts by enhancing road safety, and barely poses any negative social consequences.
    
\item {\bf Safeguards}
    \item[] Question: Does the paper describe safeguards that have been put in place for responsible release of data or models that have a high risk for misuse (e.g., pre-trained language models, image generators, or scraped datasets)?
    \item[] Answer: \answerNA{} 
    \item[] Justification: The paper does not introduce or release high-risk assets, datasets, or models requiring specific safeguards.

\item {\bf Licenses for existing assets}
    \item[] Question: Are the creators or original owners of assets (e.g., code, data, models), used in the paper, properly credited and are the license and terms of use explicitly mentioned and properly respected?
    \item[] Answer: \answerYes{} 
    \item[] Justification: The original assets (NAVSIM, nuScenes, NeuroNCAP, Qwen-VL variants) are properly cited in the references, Section 4, and Appendix B.

\item {\bf New assets}
    \item[] Question: Are new assets introduced in the paper well documented and is the documentation provided alongside the assets?
    \item[] Answer: \answerNA{} 
    \item[] Justification: The paper does not release new datasets or independent software assets.

\item {\bf Crowdsourcing and research with human subjects}
    \item[] Question: For crowdsourcing experiments and research with human subjects, does the paper include the full text of instructions given to participants and screenshots, if applicable, as well as details about compensation (if any)? 
    \item[] Answer: \answerNA{} 
    \item[] Justification: The research does not involve human subjects or crowdsourcing.

\item {\bf Institutional review board (IRB) approvals or equivalent for research with human subjects}
    \item[] Question: Does the paper describe potential risks incurred by study participants, whether such risks were disclosed to the subjects, and whether Institutional Review Board (IRB) approvals (or an equivalent approval/review based on the requirements of your country or institution) were obtained?
    \item[] Answer: \answerNA{} 
    \item[] Justification: The research does not involve human subjects or crowdsourcing.

\item {\bf Declaration of LLM usage}
    \item[] Question: Does the paper describe the usage of LLMs if it is an important, original, or non-standard component of the core methods in this research? Note that if the LLM is used only for writing, editing, or formatting purposes and does \emph{not} impact the core methodology, scientific rigor, or originality of the research, declaration is not required.
    \item[] Answer: \answerYes{} 
    \item[] Justification: The use of LLMs (Qwen3-VL-235B) to generate pseudo-labeled CoT reasoning data is explicitly documented in Section 4.1 and the exact prompts used are listed in Appendix C.

\end{enumerate}

\end{document}